\documentclass[runningheads]{llncs}
\usepackage{graphicx}
\usepackage{url}
\usepackage{xcolor}
\usepackage{amsfonts}
\usepackage{amsmath}
\usepackage{tikz}
\usepackage{subfigure}
\usepackage[moderate]{savetrees}
\usetikzlibrary{shapes.geometric, arrows}

\newcommand{\ncf}{\delta}

\newcommand{\review}[1]{#1}
\begin{document}
%
\title{Neural~Predictive~Monitoring under Partial Observability}

\author{
Francesca Cairoli\inst{1} \and
Luca Bortolussi\inst{1,2}\and
Nicola Paoletti\inst{3}
}
\authorrunning{F. Cairoli et al.}

\institute{Department of Mathematics and Geosciences, Universit\`a di Trieste, Italy \and
Modeling and Simulation Group, Saarland University, Germany
\and Department of Computer Science, Royal Holloway University, London\\}
\maketitle              
\begin{abstract}
We consider the problem of predictive monitoring (PM), i.e., predicting at runtime future violations of a system from the current state. We work under the most realistic settings where only partial and noisy observations of the state are available at runtime. Such settings directly affect the accuracy and reliability of the reachability predictions, jeopardizing the safety of the system. 
In this work, we present a learning-based method for PM that produces accurate and reliable reachability predictions despite partial observability (PO).
We build on Neural Predictive Monitoring (NPM), a PM method that uses deep neural networks for approximating hybrid systems reachability, and extend it to the PO case. We propose and compare two solutions, an \textit{end-to-end} approach, which directly operates on the rough observations, and a \textit{two-step} approach, which introduces an intermediate state estimation step. Both solutions rely on conformal prediction to provide 1) probabilistic guarantees in the form of prediction regions and 2) sound estimates of predictive uncertainty. We use the latter to identify unreliable (and likely erroneous) predictions and to retrain and improve the monitors on these uncertain inputs (i.e., active learning). Our method results in highly accurate reachability predictions and error detection, as well as tight prediction regions with guaranteed coverage. 

\end{abstract}

\section{Introduction}\label{sec:intro}
We focus on \textit{predictive monitoring (PM) of cyber-physical systems (CPSs)}, that is, the problem of predicting, at runtime, if a safety violation is imminent from the current CPS state. In particular, we work under the (common) setting where the true CPS state is unknown and we  only can access partial (and noisy) observations of the system. 

With CPSs having become ubiquitous in safety-critical domains, from autonomous vehicles to medical devices~\cite{alur2015principles}, runtime safety assurance of these systems is paramount. In this context, PM has the advantage, compared to traditional monitoring~\cite{bartocci2018specification}, of detecting potential safety violations before they occur, in this way enabling preemptive  countermeasures to steer the system back to safety (e.g., switching to a failsafe mode as done in the Simplex architecture~\cite{johnson2016real}). 
Thus, effective PM must balance between prediction accuracy, to avoid errors that can jeopardize safety, and computational efficiency, to support fast execution at runtime. Partial observability (PO) makes the problem more challenging, as it requires some form of state estimation (SE) to reconstruct the CPS state from observations: 
on top of its computational overhead, SE introduces estimation errors that propagate in the reachability predictions, affecting the PM reliability. Existing PM approaches either assume full state observability~\cite{bortolussi2019neural} or cannot provide correctness guarantees on the combined estimation-prediction process~\cite{chou2020predictive}. 

We present a learning-based method for predictive monitoring designed to produce efficient and highly reliable reachability predictions  under noise and partial observability. We build on neural predictive monitoring (NPM)~\cite{bortolussi2019neural,bortolussineural}, an approach that employs neural network classifiers to predict reachability at any given state. 
Such an approach is both accurate, owing to the expressiveness of neural networks (which can approximate well hybrid systems reachability given sufficient training data~\cite{phan2018neural}), and efficient, since the  analysis at runtime boils down to a simple forward pass of the neural network. 

\begin{figure}
    \centering
    \includegraphics[scale=0.8]{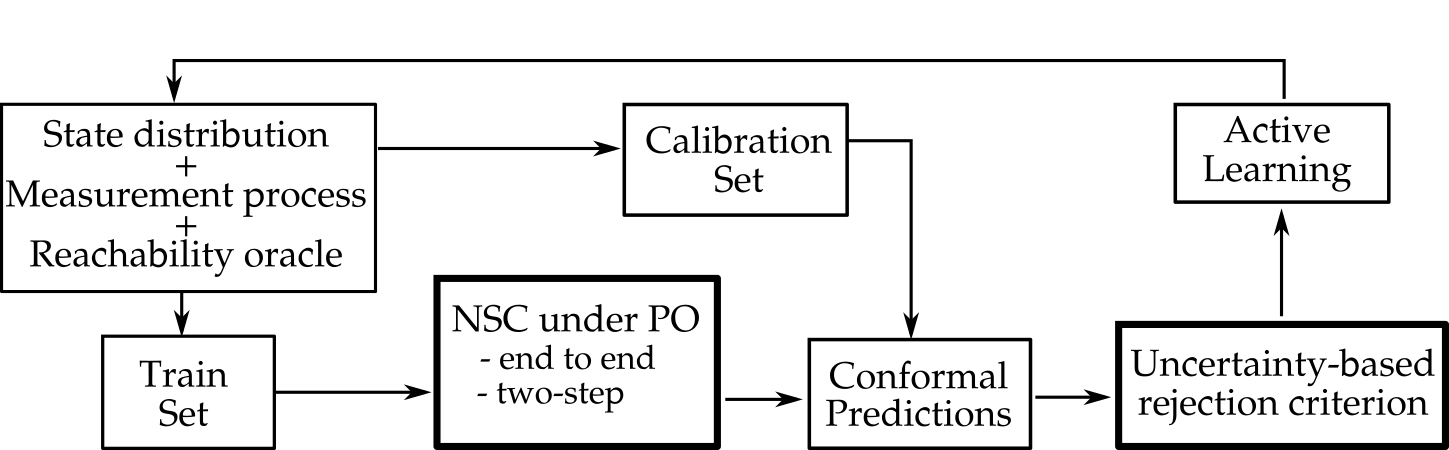}
    \caption{Overview of the NPM framework under partial observability. The components used at runtime have a thicker border. \vspace{-0.3cm}}
    \label{fig:diagram}
\end{figure}

We extend and generalize NPM to the PO setting by investigating two solution strategies: an \textit{end-to-end} approach where the neural monitor directly operates on the raw observations (i.e., without reconstructing the state); and a \textit{two-step} approach, where it operates on state sequences estimated from observations using a dedicated neural network model. See Fig \ref{fig:diagram} for an overview of the approach.

Independently of the strategy chosen for handling PO, our approach offers two ways of quantifying and enhancing PM reliability. Both are based on conformal prediction~\cite{balasubramanian2014conformal,vovk2005algorithmic}, a popular framework for reliable machine learning. First, we complement the predictions of the neural monitor and state estimator with prediction regions guaranteed to cover the true (unknown) value with arbitrary probability. To our knowledge, we are the first to provide probabilistic guarantees on state estimation and reachability under PO. Second, as in NPM, we use measures of predictive uncertainty to derive optimal criteria for detecting (and rejecting) potentially erroneous predictions. These rejection criteria also enable active learning, i.e., retraining and improving the monitor on such identified uncertain predictions. 

We evaluate our method on a benchmark of six hybrid system models. Despite PO, we obtain highly accurate reachability predictions (with accuracy above 99\% for most case studies). These results are further improved by our uncertainty-based rejection criteria, which manage to preemptively identify the majority of prediction errors (with a detection rate close to 100\% for most models). In particular, we find that the two-step approach tends to outperform the end-to-end one. The former indeed benefits from a neural SE model, which provides high-quality state reconstructions and is empirically superior to Kalman filters~\cite{wan2000unscented} and moving horizon estimation~\cite{allgower1999nonlinear},
two of the main SE methods. Moreover, our method produces prediction regions that are efficient (i.e., tight) yet satisfy the \textit{a priori} guarantees. Finally, we show that active learning not just improves reachability prediction and error detection, but also increases both coverage and efficiency of the prediction regions, which implies stronger guarantees and less conservative regions. 

\section{Problem Statement}\label{sec:problem}


We consider hybrid systems (HS) with discrete time and deterministic dynamics and state space $S = V\times Q$, where $V\subseteq \mathbb{R}^n$ is the domain of the continuous variables, and $Q$ is the set of discrete modes. 
\begin{equation}\label{eq:dynamics}
    v_{i+1} =F_{q_i}(v_i, a_i, t_i); \quad  
    q_{i+1} = J_{q_i}(v_i); \quad
    a_i = C_{q_i}(v_i); \quad
    y_i = \mu(v_i,q_i)+w_i,
\end{equation}

where $v_i = v(t_i)$, $q_i = q(t_i)$, $a_i = a(t_i)$, $y_i = y(t_i)$ and $t_i = t_0+i\cdot\Delta t$. 
Given a mode $q\in Q$, 
$F_q$ 
is the mode-dependent dynamics of the continuous component, $J_q$ is mode switches (i.e., discrete jumps),
$C_q$ is the (given) control law.  
Partial and noisy observations $y_i \in Y$ are produced by the observation function $\mu$ and the additive measurement noise $w_i\sim \mathcal{W}$ (e.g., white Gaussian noise). 



Predictive monitoring of such a system corresponds to deriving a function that approximates a given reachability specification  $\mathsf{Reach}(U,s,H_f)$: given a state $s=(v,q)$ and a set of unsafe states $U$, establish whether the HS admit a trajectory starting from $s$ that reaches $U$ in a time $H_f$. The approximation is w.r.t.\ some given distribution of HS states, meaning that we can admit inaccurate reachability predictions if the state has zero probability. 
We now illustrate the PM problem under the ideal assumption that the full HS can be accessed.

\begin{problem}[PM for HS under full observability]\label{prbl:pm}
Given an HS~\eqref{eq:dynamics} with state space $S$, a distribution $\mathcal{S}$ over $S$, a time bound $H_f$ and set of unsafe states $U \subset S$, find a function $h^*: S\rightarrow \{0,1\}$ that minimizes the probability
\[
Pr_{s\sim\mathcal{S}}\Big(
h^*(s)\ne\mathbf{1}\big(\mathsf{Reach}(U,s,H_f)\big)
\Big),
\]
where $\mathbf{1}$ is the indicator function. 
A state $s \in S$ is called \emph{positive} w.r.t a predictor $h: S\rightarrow \{0,1\}$ if $h(s) = 1$.  
Otherwise, $s$ is called \emph{negative}. 
\end{problem}

As discussed in the next section, finding $h^*$, i.e., finding a function approximation with minimal error probability, can be solved as a supervised classification problem, provided that a reachability oracle is available for generating supervision data. 

The problem above relies on the assumption that full knowledge about the HS state is available. However, in most practical applications, state information is partial and noisy. 
Under PO, we only have access to a sequence of past observations $\mathbf{y}_t = (y_{t-H_p},\dots,y_t)$ which are generated as per~\eqref{eq:dynamics}, that is, by applying the observation function $\mu$ and measurement noise to the \textit{unknown} state sequence $s_{t-H_p},\dots,s_t$. 

In the following, we consider the distribution $\mathcal{Y}$ over $Y^{H_p}$ of the observations sequences $\mathbf{y}_t = (y_{t-H_p},\dots,y_t)$ induced by state  $s_{t-H_p}\sim\mathcal{S}$, HS dynamics~\eqref{eq:dynamics}, and iid noise $\mathbf{w}_t = (w_{t-H_p},\dots,w_t) \sim \mathcal{W}^{H_p}$. 


\begin{problem}[PM for HS under noise and partial observability]\label{prbl:pm_po}
Given the HS and reachability specification of Problem~\ref{prbl:pm}, 
find a function $g^*: Y^{H_p} \rightarrow \{0,1\}$ that minimizes 
\[{Pr}_{\mathbf{y}_t \sim \mathcal{Y}}
\Big(
g^*\big(\mathbf{y}_t\big)\ne\mathbf{1}\big( \mathsf{Reach}(U,s_t,H_f)\big)
\Big).
\]
\end{problem}



In other words, $g^*$ should predict reachability values given in input only a sequence of past observations, instead of the true HS state. 
In particular, we require a sequence of observations for the sake of identifiability. Indeed, for general non linear systems, a single observation does not contain enough information to infer the HS state\footnote{Feasibility of state reconstruction is affected by the time lag and the sequence length. Our focus is to derive the best predictions for fixed lag and sequence length, not to fine-tune these to improve identifiability.}.

The predictor $g$ is an approximate solution and, as such, it can commit safety-critical prediction errors. Building on~\cite{bortolussi2019neural}, we endow the predictive monitor with an  
error detection criterion $R$. This criterion should be able to \textit{preemptively} identify -- and hence, reject -- sequences of observations $\mathbf{y}$ where $g$'s prediction is likely to be erroneous (in which case $R$ evaluates to $1$, $0$ otherwise). $R$ should also be optimal in that it has minimal probability of detection errors. The rationale behind $R$ is that uncertain predictions are more likely to lead to prediction errors. Hence, rather than operating directly over observations $\mathbf{y}$, the detector $R$ receives in input a measure of predictive uncertainty of $g$ about $\mathbf{y}$. 

\begin{problem}[Uncertainty-based error detection under noise and partial observability]\label{prbl:rejection}
Given an approximate reachability predictor $g$ for the HS and reachability specification of Problem~\ref{prbl:pm_po}, and a measure of predictive uncertainty $u_g: Y^{H_p}\rightarrow D$ over some uncertainty domain $D$, find an optimal error detection rule, $R^*_{g}:D\rightarrow \{0,1\}$, that minimizes the probability
$$
Pr_{\mathbf{y_t}\sim \mathcal{Y}} \ \mathbf{1}\Big(g(\mathbf{y_t}) \neq \mathbf{1}(\mathsf{Reach}(U,s_t,H_f)) \Big) \neq R^*_{g}(u_g(\mathbf{y_t})).
$$ 
\end{problem}
In the above problem, we consider all kinds of prediction errors, but the definition and approach could be easily adapted to focus on the detection of only e.g., false negatives (the most problematic errors from a safety-critical viewpoint). 

The general goal of Problems \ref{prbl:pm_po} and \ref{prbl:rejection} is to minimize the risk of making mistakes in predicting reachability and predicting predictions errors, respectively. We are also interested in establishing probabilistic guarantees on the expected error rate, in the form of predictions regions guaranteed to include the true reachability value with arbitrary probability.

\begin{problem}[Probabilistic guarantees]\label{prbl:stat_guar}
Given the HS and reachability specification of Problem~\ref{prbl:pm_po}, 
find a function $\Gamma^{\epsilon}: Y^{H_p}\rightarrow 2^{\{0,1\}}$, mapping a sequence of past observations $\mathbf{y}$ into a prediction region for the corresponding reachability value, i.e., a region that satisfies, for any error probability level $\epsilon \in (0,1)$, the \textit{validity} property below
\[
Pr_{\mathbf{y_t}\sim \mathcal{Y}}
\Big(
\mathbf{1}\big(\mathsf{Reach}(U,s_t,H_f)\big)\in \Gamma^{\epsilon}\big(\mathbf{y}_t\big)
\Big)\ge 1-\epsilon.
\]
\end{problem}
Among the maps that satisfy validity, we seek the most \emph{efficient} one, meaning the one with the smallest, i.e. less conservative, prediction regions. 

\section{Methods}\label{sec:new_methods}

In this section, we first describe our learning-based solution to PM under PO (Problem~\ref{prbl:pm_po}). 
We then provide background on conformal prediction (CP) and explain how we apply this technique to endow our reachability predictions and state estimates with probabilistic guarantees (Problem~\ref{prbl:stat_guar}). Finally, we illustrate how CP can be used to derive measures of predictive uncertainty to enable error detection (Problem~\ref{prbl:rejection}) and active learning. 

\subsection{Predictive Monitoring under Noise and Partial Observability}\label{sec:pm_po}

There are two natural learning-based approaches to tackle Problem~\ref{prbl:pm_po} (see Fig.~\ref{fig:nsc_diagram}):
\begin{enumerate}
    \item an \textbf{end-to-end} solution that learns a direct mapping from the sequence of past measurements $\mathbf{y}_t$ to the reachability label $\{0,1\}$.
    
    \item a \textbf{two-step} solution that combines steps (a) and (b) below:  
    \begin{itemize}
        \item[(a)] learns a \textit{state estimator} able to reconstruct the history of full states $\mathbf{s}_t = (s_{t-H_p},\dots,s_t)$ from the sequence of measurements $\mathbf{y}_t = (y_{t-H_p},\dots,y_t)$;
        \item[(b)] learns a \textit{state classifier} mapping the sequence of states $\mathbf{s}_t$ to the reachability label $\{0,1\}$;
    \end{itemize}
\end{enumerate}

\subsubsection{Dataset Generation.} Since we aim to solve the PM problem as one of supervised learning, the first step is generating a suitable training dataset. 
For this purpose, we need reachability oracles to label states $s$ as safe (negative), if $\neg \mathsf{Reach}(U,s,H_f)$, or unsafe (positive) otherwise. Given that we consider deterministic HS dynamics, we use simulation (rather than reachability checkers like~\cite{chen2013flow,althoff2016implementation,bogomolov2019juliareach}) to label the states. 

The reachability of the system at time $t$ depends only on the state of the system at time $t$, however, one can decide to exploit more information and make a prediction based on the previous $H_p$ states. 
Formally, the generated dataset under full observability can be expressed as 
$\mathcal{D}_{NPM} = \{(\mathbf{s}^i_t, l^i)\}_{i=1}^N$, where $\mathbf{s}^i_t = (s^i_{t-H_p},s^i_{t-H_p+1},\dots, s^i_t)$ and $l^i = \mathbf{1}(\mathsf{Reach}(U,s^i_t,H_f))$
. 
Under partial observability, we use the (known) observation function $\mu:S\rightarrow Y$ to build a dataset 
$\mathcal{D}_{PO-NPM}$ made of tuples $(\mathbf{y}_t, \mathbf{s}_t, l_t)$, where $\mathbf{y}_t$ is a sequence of noisy observations for $\mathbf{s}_t$, i.e., such that $\forall j\in\{t-H_p,\dots , t\}$ $y_{j} = \mu(s_j)+w_j$ and $w_j\sim\mathcal{W}$.
The distribution of $\mathbf{s}_t$ and $\mathbf{y}_t$ is determined by the distribution $\mathcal{S}$ of the initial state of the sequences, $s_{t-H_p}$. 

We consider two different distributions:  \emph{independent}, where the initial states $s_{t-H_p}$ are sampled independently, thus resulting in independent state/observation sequences; and 
\emph{sequential}, where states come from temporally correlated trajectories in a sliding-window fashion. The latter is more suitable for real-world runtime applications, where observations are received in a sequential manner. On the other hand, temporal dependency violates the exchangeability property, which affects the theoretical validity guarantees of CP, as we will soon discuss.

\tikzstyle{arrow} = [thick,->,>=stealth]
\tikzstyle{rect} = [rectangle, rounded corners, minimum width=3cm, minimum height=1cm,text centered, draw=black]

\begin{figure}[!t]
    \centering
    \resizebox{0.7\textwidth}{!}{%
\begin{tikzpicture}

\node (meas) [rect] {$\mathbf{y}_t = ( y_{t-H_p},\dots , y_t) $};

\node (label) [rect, right of=meas, xshift=5cm] {$\mathsf{Reach}(U,s_t,H_f)$};

\draw [arrow] (meas) -- node[anchor=south] {(1) end-to-end} (label);

\node (state) [rect, below of= meas, yshift=2.75cm] {$\mathbf{s}_t = (s_{t-H_p},\dots , s_t)$};

\draw [arrow] (state) -- node[anchor=west] {\qquad (2.b) state-classifier} (label);

\draw [arrow] (meas) -- node[anchor=east] {(2.a) state-estimator} (state);
\end{tikzpicture}
}%
    \caption{Diagram of NSC under noise and partial observability. }
    \label{fig:nsc_diagram}
\end{figure}
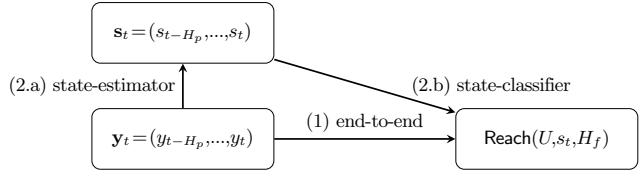

Starting from $\mathcal{D}_{PO-NPM}$, the two alternative approaches, end-to-end and two-step, can be developed as follows.

\subsubsection{End-to-end solution.}

We train a one-dimensional convolutional neural net (CNN) that learns a direct mapping from $\mathbf{y}_t$ to $l_t$, i.e., we solve a simple binary classification problem. This approach ignores the sequence of states $\mathbf{s}_t$. The canonical binary cross-entropy function can be considered as loss function for the weights optimization process.

\subsubsection{Two-step solution.}
A CNN regressor, referred to as Neural State Estimator (NSE), is trained to reconstruct the sequence of states $\hat{\mathbf{s}}_t$ from the sequence of noisy observations $\mathbf{y}_t$. This is combined with, a CNN classifier, referred to as Neural State Classifier (NSC), trained to predict the reachability label $l_t$ from the sequence of states $\mathbf{s}_t$. The mean square error between the sequences of real states $\mathbf{s}_t$ and the reconstructed ones $\hat{\mathbf{s}}_t$ is a suitable loss function for the NSE, whereas for the NSC we use, once again, a binary cross-entropy function. 

The network resulting from the combination of the the NSE and the NSC maps the sequence of noisy measurements into the safety label, exactly as required in Problem~\ref{prbl:pm_po}. However, the NSE inevitably introduces some errors in reconstructing $\mathbf{s}_t$. Such error is then propagated when the NSC is evaluated on the reconstructed state, $\hat{\mathbf{s}}_t$, as it is generated from a distribution different from $\mathcal{S}$, affecting the overall accuracy of the combined net. 
To alleviate this problem, we introduce a \emph{fine-tuning} phase in which the weights of the NSE and the weights of the NSC are updated together, minimizing the sum of the two respective loss functions. 
In this phase, the NSC learns to classify correctly the state reconstructed by the NSE, $\hat{\mathbf{s}}_t$, rather than the real state $\mathbf{s}_t$, so to improve the task specific accuracy. 

\paragraph{Neural State Estimation.}
The two-step approach has an important additional advantage, the NSE. In general, any traditional state estimator could have been used. Nevertheless, non-linear systems make SE extremely challenging for existing approaches. On the contrary, our NSE reaches very high reconstruction precision (as demonstrated in the result section). Furthermore, because of the fine-tuning, it is possible to calibrate the estimates to be more accurate in regions of the state-space that are safety-critical.



\subsection{Conformal Prediction for regression and classification}\label{sec:cp}


In the following, \review{we provide background on conformal prediction considering} 
a generic prediction model. Let $X$ be the input space, $T$ be the target space, and define $Z = X\times T$. Let $\mathcal{Z}$ be the data-generating distribution, i.e., the distribution of the points $(x,t)\in Z$.  
The prediction model is represented as a function $f:X\rightarrow T$. 
For a generic input $x$, 
we denote with $t$ the true target value of $x$ and with $\hat{t}$ the prediction by $f$. 
Test inputs, whose unknown true target values we aim to predict, are denoted by $x_*$.

In our setting of reachability prediction, inputs are observation sequences, target values are the corresponding reachability values. The data distribution $\mathcal{Z}$ is the joint distribution of observation sequences and  reachability values induced by state $s_{t-H_P}\sim \mathcal{S}$ and iid noise vector $\mathbf{w}_t\sim\mathcal{W}^{H_p}$.

Conformal Prediction associates measures of reliability to any traditional supervised learning problem. It is a very general approach that can be applied across all existing classification and regression methods~\cite{balasubramanian2014conformal,vovk2005algorithmic}. 
CP produces \textit{prediction regions with guaranteed validity}, thus satisfying the statistical guarantees illustrated in Problem~\ref{prbl:stat_guar}.

\begin{definition}[Prediction region]
For significance level $\epsilon \in (0,1)$ and test input $x_*$, the $\epsilon$-prediction region for $x_*$, $\Gamma_*^{\epsilon}\subseteq T$, is a set of target values s.t.
\begin{equation}\label{eq:pred_r}
    \underset{(x_*,t_*)\sim \mathcal{Z}}{Pr}(t_* \in \Gamma_*^{\epsilon}) = 1 - \epsilon.
\end{equation}
\end{definition}

The idea of CP is to construct the prediction region by ``inverting'' a suitable hypothesis test: given a test point $x_*$ and a tentative target value $t'$, we \textit{exclude} $t'$ from the prediction region only if it is unlikely that $t'$ is the true value for $x_*$. The test statistic is given by a so-called \textit{nonconformity function (NCF)} $\ncf:Z \rightarrow \mathbb{R}$, which, given a predictor $f$ and a point $z=(x,t)$, measures the deviation between the true value $t$ and the corresponding prediction $f(x)$.  In this sense, $\ncf$ can be viewed as a generalized residual function. In other words, CP builds the prediction region $\Gamma_*^{\epsilon}$ for a test point $x_*$ by excluding all targets $t'$ whose NCF values are unlikely to follow the NCF distribution of the true targets:
\begin{equation}\label{eq:cp_predr}
\Gamma_*^{\epsilon} = \left\{t' \in T \mid  Pr_{(x,t)\sim \mathcal{Z}}\left(\ncf(x_*,t') \geq \ncf(x,t)\right) > \epsilon\right\}.
\end{equation}
The probability term in Eq.~\ref{eq:cp_predr} is often called p-value. 
From a practical viewpoint, the NCF distribution $Pr_{(x,t)\sim \mathcal{Z}}(\ncf(x,t))$ cannot be derived in an analytical form, and thus we use an empirical approximation derived using a sample $Z_c$ of $\mathcal{Z}$.  This  approach is called \textit{inductive CP}~\cite{papadopoulos2008inductive} and $Z_c$ is referred to as \textit{calibration set}.

\begin{remark}[Assumptions and guarantees of inductive CP] Importantly, CP prediction regions have \textit{finite-sample validity}~\cite{balasubramanian2014conformal}, i.e., they satisfy~\eqref{eq:pred_r} for any sample of $\mathcal{Z}$ (or reasonable size), and not just asymptotically. On the other hand, CP's theoretical guarantees hold under the \textit{exchangeability} assumption (a ``relaxed'' version of iid) by which the joint probability of any sample of $\mathcal{Z}$ is invariant to permutations of the sampled points. Of the two observation distributions discussed in Section~\ref{sec:problem}, we have that independent observations are exchangeable but sequential ones are not (due to the temporal dependency). Even though sequential data violate CP's theoretical validity, we find that the prediction regions still attain empirical coverage consistent with the nominal coverage (see results section), that is, the probabilistic guarantees still hold in practice (as also found in previous work on CP and time-series data~\cite{balasubramanian2014conformal}).
\end{remark}

\paragraph{Validity and Efficiency.} 
CP performance is measured via two quantities: 1) \emph{validity} (or \emph{coverage}), i.e. the empirical error rate observed on a test sample, which should be as close as possible to the significance level $\epsilon$, and 2) \emph{efficiency}, i.e. the size of the prediction regions, which should be small. CP-based prediction regions are automatically valid (under the assumptions of Remark 1), whereas the efficiency depends on the chosen nonconformity function and thus the underlying model.

\subsubsection{CP for classification.} In classification, the target space is a discrete set of possible labels (or classes) $T=\{\ell^1,\ldots,\ell^c\}$.  We represent the classification model as a function $f:X\rightarrow [0,1]^c$ mapping inputs into a vector of class likelihoods, such that the predicted class is the one with the highest likelihood\footnote{Ties can be resolved by imposing an ordering over the classes.}. 
Classification is relevant for predictive monitoring as the reachability predictor of Problem~\ref{prbl:pm_po} is indeed a binary classifier ($T=\{0,1\}$) telling whether or not an unsafe state can be reached given a sequence of observation. 

The inductive CP algorithm for classification is divided into an offline phase, executed only once, and an online phase, executed for every test point $x_*$. In the offline phase (steps 1--3 below), we train the classifier $f$ and construct the calibration distribution, i.e., the empirical approximation of the NCF distribution. In the online phase (steps 4--5), we derive the prediction region for $x_*$ using the computed classifier and distribution.
\begin{enumerate}
    \item Draw sample $Z'$ of $\mathcal{Z}$. Split $Z'$ into training set $Z_t$ and calibration set $Z_c$.
    \item Train classifier $f$ using $Z_t$. Use $f$ to define an NCF $\delta$.
    \item Construct the calibration distribution by computing, for each $z_i \in Z_c$, the NCF score $\alpha_i = \delta(z_i)$.
    \item For each label $\ell^j \in T$, compute $\alpha_*^j = \delta(x_*,\ell^j)$, i.e., the NCF score for $x_*$ and $\ell^j$, and the associated p-value $p_*^{j}$:
    \begin{equation}\label{eq:smoothed_p}
p_*^{j}= \frac{|\{z_i\in Z_c \mid \alpha_i > \alpha_*^{j}\}|}{|Z_c|+1}+\theta\frac{|\{z_i\in Z_c \mid \alpha_i = \alpha_*^{j}\}|+1}{|Z_c|+1},
\end{equation}
where $\theta\in\mathcal{U}[0,1]$ is a tie-breaking random variable. 
\item Return the prediction region $\Gamma_*^{\epsilon} = \{\ell^j\in T \mid p_*^{j}>\epsilon\}.$
\end{enumerate}
In defining the NCF $\delta$, we should aim to obtain high $\delta$ values for wrong predictions and low $\delta$ values for correct ones. Thus, a natural choice in classification is to define $\delta(x,l^j) = 1 - f(x)_j$, where $f(x)_j$ is the likelihood predicted by $f$ for class $l_j$. Indeed, if $l^j$ is the true target for $x$ and $f$ correctly predicts $l^j$, then $f(x)_j$ is high (the highest among all classes) and $\delta(x,l^j)$ is low; the opposite holds if $f$ does not predict $l^j$.

\subsubsection{CP for Regression.}\label{sec:cp_regr}
In regression we have a continuous target space $T\subseteq\mathbb{R}^n$. Thus, the regression case is relevant for us because our state estimator can be viewed as a regression model, where $T$ is the state space. 

The CP algorithm for regression is similar to the classification one. In particular, the offline phase of steps 1--3, i.e., training of regression model $f$ and definition of NCF $\delta$, is the same (with obviously a different kind of $f$ and $\delta$). 

The online phase changes though, because $T$ is a continuous space and thus, it is not possible to enumerate the target values and compute for each a p-value. Instead, we proceed in an equivalent manner, that is, identify the critical value $\alpha_{(\epsilon)}$ of the calibration distribution, i.e., the NCF score corresponding to a p-value of $\epsilon$. The resulting $\epsilon$-prediction region is given by $\Gamma_*^{\epsilon} = f(x_*) \pm \alpha_{(\epsilon)}$, where $\alpha_{(\epsilon)}$ is the $(1-\epsilon)$-quantile of the calibration distribution, i.e., the $\lfloor \epsilon \cdot (|Z_c|+1)\rfloor$-th largest calibration score\footnote{Such  prediction intervals have the same width ($\alpha_{(\epsilon)}$) for all inputs. There are techniques like~\cite{romano2019conformalized} that allow to construct intervals with input-dependent widths, which can be equivalently applied to our problem.}. 

A natural NCF in regression, and the one used in our experiments, is the norm of the difference between the real and the predicted target value, i.e., $\delta(x) = ||t - f(x)||$. 





\subsection{CP-based quantification of predictive uncertainty}
We illustrate how to complement reachability predictions with uncertainty-based error detection rules, which leverage measures of predictive uncertainty to preemptively identify the occurrence of  prediction errors. 
Detecting errors efficiently requires a fine balance between the number of errors accurately prevented and the overall number of discarded predictions.

We use two uncertainty measures, \emph{confidence} and \emph{credibility}, that are extracted from the CP algorithm for classification. The method discussed below was first introduced for NPM~\cite{bortolussi2019neural}, but here this is extended to the PO case.

\subsubsection{Confidence and credibility.}\label{sec:conf_cred}
Let us start by observing that, for significance levels  $\epsilon_1\ge\epsilon_2$, the corresponding prediction regions are such that $\Gamma^{\epsilon_1}\subseteq \Gamma^{\epsilon_2}$.  
It follows that, given an input $x_*$, if $\epsilon$ is lower than all its p-values, i.e. $\epsilon < \min_{j=1,\ldots,c} \ p_*^{j}$, then the region $\Gamma_*^{\epsilon}$ contains all the labels. As $\epsilon$ increases, fewer and fewer classes will have a p-value higher than $\epsilon$. That is, the region shrinks as $\epsilon$ increases. In particular, $\Gamma_*^{\epsilon}$ is empty when $\epsilon \geq \max_{j=1,\ldots,c} \ p_*^{j}$.  

The \textit{confidence} of a point $x_*\in X$, $1-\gamma_*$, measures how likely is our prediction for $x_*$ compared to all other possible classifications (according to the calibration set). It is computed as one minus the smallest value of $\epsilon$ for which the conformal region is a single label, i.e. the second largest p-value $\gamma_*$: 
\[1-\gamma_* = \sup \{1-\epsilon : |\Gamma_*^{\epsilon}| = 1\}.\]

The \textit{credibility}, $c_*$, indicates how suitable the training data are to classify that specific example. In practice, it is the smallest $\epsilon$ for which the prediction region is empty, i.e. the highest p-value according to the calibration set, which corresponds to the p-value of the predicted class: 
\[c_* = \inf \{\epsilon : |\Gamma_*^{\epsilon}|= 0\}.\]

Note that if $\gamma_*\le \epsilon$, then the corresponding prediction region $\Gamma_*^{\epsilon}$ contains at most one class. If both $\gamma_*\le \epsilon$ and $c_* > \epsilon$ hold, then the prediction region contains \textit{exactly} one class, denoted as $\hat{\ell}_*$, i.e., the one predicted by $f$. In other words, the interval $[\gamma_*, c_*)$ contains all the $\epsilon$ values for which we are sure that $\Gamma_*^{\epsilon}=\{\hat{\ell}_*\}$. 
It follows that the higher $1-\gamma_*$ and $c_*$ are, the more reliable the prediction $\hat{\ell}_*$ is, because we have an expanded range $[\gamma_*, c_*)$ of significance values by which $\hat{\ell}_*$ is valid. 
Indeed, in the extreme scenario where $c_*=1$ and $\gamma_*=0$, then $\Gamma_*^{\epsilon}=\{\hat{\ell}_*\}$ for any value of $\epsilon$. This is why, as we will soon explain, our uncertainty-based rejection criterion relies on excluding points with low values of $1-\gamma_*$ and $c_*$. 
In binary classification problems, each point $x_*$ has only two p-values, one for each class, which coincide with $c_*$ (p-value of the predicted class) and $\gamma_*$ (p-value of the other class).

Given a reachability predictor $g$, the uncertainty function $u_g$ can be defined as the function mapping a sequence of observations $\mathbf{y}^*$ into the confidence $\gamma^*$ and the credibility $c^*$ of $g(\mathbf{y}^*)$, thus $u_g(\mathbf{y}^*) = (\gamma^*, c^*)$. 
In order to learn a good decision rule to identify trustworthy predictions, 
we solve another binary classification problem on the  uncertainty values.
In particular, we use a cross-validation strategy to compute values of confidence and credibility over the entire calibration set, as it is not used to train the classifier, and label each point as $0$ if it is correctly classified by the predictor and as $1$ if it is misclassified. We then train a Support Vector Classifier (SVC) that automatically learns to distinguish points that are misclassified from points that are correctly classified based on the values of confidence and credibility. \review{In particular, we choose a simple linear classifier as it turns out to perform satisfactorily well, especially on strongly unbalanced datasets. 
Nevertheless, other kinds of classifiers can be applied as well.}

To summarize, given a predictor $g$ and a new sequence of observations $\mathbf{y}^*$, we obtain a prediction about its safety, $g(\mathbf{y}^*) = \hat{l}^*$, and a quantification of its uncertainty, $u^* = u_g(\mathbf{y}^*) = (\gamma^*,c^*)$. If we feed $u^*$ to the rejection rule $R_g$ we obtain a prediction about whether or not the prediction of $g$ about $\mathbf{y}^*$ can be trusted.

\subsection{Active Learning (AL) }

NPM depends on two related learning problems: the reachabiliy predictor $g$ and the rejection rule $R_g$. 
We leverage the \textit{uncertainty-aware active learning} solution presented in~\cite{bortolussineural}, where the re-training points are derived by first sampling a large pool of unlabeled data, and then considering only those points where the current predictor $g$ is still uncertain, i.e. those points which are rejected by our rejection rule $R_g$.
A fraction of the labeled samples is added to the training set, whereas the remaining part is added to the calibration set, keeping the training/calibration ratio constant. As a matter of fact, a principled criterion to select the most informative samples would benefit both the accuracy and the efficiency of the method, as the size of the calibration set affects the runtime efficiency of the error detection rule. 

The addition of such actively selected points results in a shift of the data generating distribution, that does not match anymore the distribution of the test samples. This implies that the theoretical guarantees of CP are lost. However, as we will show in the experiments, AL typically results in an empirical increase of the coverage, i.e., in even stronger probabilistic guarantees. The reason is that AL is designed to improve on poor predictions, which, as such, have prediction regions more likely to miss the true value. Improving such poor predictions thus directly cause an increased coverage (assuming that the classifier remains accurate enough on the inputs prior to AL).


\section{Experimental Evaluation}\label{sec:experiments}

We evaluate both end-to-end and two-step approaches under PO on six benchmarks of cyber-physical systems with dynamics presenting a varying degree of complexity and with a variety of observation functions. 
We include white Gaussian noise to introduce stochasticity in the observations.

\subsection{Case Studies}

\begin{itemize}
    \item \textbf{IP}: classic two-dimensional non-linear model of an Inverted Pendulum on a cart. Given a state $s=(s_1,s_2)$, we observe a noisy measure of the energy of the system $y = s_2/2+\cos(s_1)-1 + w$, where $w\sim\mathcal{N}(0,0.005)$. Unsafe region $U=\{s:  |s_1|\ge\pi/6\}$. $H_p = 1$, $H_f = 5$.
    \item \textbf{SN}: a two-dimensional non-linear model of the Spiking Neuron action potential. Given a state $s=(s_1,s_2)$ we observe a noisy measure of $s_2$, $y= s_2+w$, with $w\sim\mathcal{N}(0,0.1)$. Unsafe region $ U = \{s: s_1 \le -68.5\}$. $H_p = 4$, $H_f = 16$.
    \item \textbf{CVDP}: a four-dimensional non-linear model of the Coupled  Van Der  Pol oscillator~\cite{ernst2020arch}, modeling two coupled oscillators. Given a state $s=(s_1,s_2,s_3,s_4)$ we observe $y = (s_1,s_3)+ w$, with $w\sim\mathcal{N}(\underbar{0},0.01\cdot I_2)$. Unsafe region $ U = \{s: s_{2} \ge 2.75 \land s_{2} \ge 2.75\}$. $H_p = 8$, $H_f = 7$.
    \item\textbf{LALO}: the seven-dimensional non-linear Laub Loomis model~\cite{ernst2020arch} of a class of enzymatic activities. Given a state $s = (s_1,s_2, s_3, s_4, s_5, s_6, s_7)$ we observe $y = (s_1, s_2, s_3, s_5, s_6, s_7)+w$, with $w\sim \mathcal{N}(\underbar{0},0.01\cdot I_6)$. Unsafe region $ U = \{s: s_4\ge 4.5\}$. $H_p = 5$, $H_f = 20$.
    \item \textbf{TWT}: a three-dimensional non-linear model of a Triple Water Tank. Given a state $s= (s_1,s_2,s_3)$ we observe $y= s+w$, with $w\sim\mathcal{N}(\underbar{0},0.01\cdot I_3)$. Unsafe region $U= \{s: \lor_{i=1}^3 s_i\not\in [4.5,5.5]\}$. $H_p = 1$, $H_f = 1$.
    \item \textbf{HC}: the 28-dimensional linear model of an Helicopter controller. We observe only the altitude, i.e. $y = s_8 + w$, with $w\sim\mathcal{N}(0,1)$. Unsafe region $U=\{s:s_8<0\}$. $H_p = 5$, $H_f = 5$.
\end{itemize}
Details about the case studies are available in the Appendix~\ref{appendix:models}. 

\subsection{Experimental settings.} 

\paragraph{Implementation.} The workflow can be divided in steps: (1) define the CPS models, (2)  generate  the synthetic datasets $\mathcal{D}_{PO-NPM}$ (both the independent and the sequential version), (3) train the NPM (either end-to-end or two-step), (4) train the CP-based error detection rules, (5) perform active learning and (6) evaluate  both the initial and the active NPM on a test set. 
From here on, we call \emph{initial setting} the one with no active learning involved.
The technique is fully implemented in Python\footnote{The experiments were performed on a computer with a CPU Intel x86, 24 cores and a 128GB RAM and 15GB of GPU Tesla V100.}. In particular, PyTorch~\cite{paszke2017automatic} is used to craft, train and evaluate the desired CNN architectures. Details about the CNN architectures and  the settings of the optimization algorithm are described in Appendix~\ref{sec:arch}.
The source code for all the experiments can be found at the following link:
\url{https://github.com/francescacairoli/Stoch_NSC.git}

\paragraph{Datasets.}

For each case study we generate both an independent and a sequential dataset. 
\begin{itemize}
    \item \textit{Independent: } the train set consists of 50K independent sequences of states of length $32$, the respective noisy measurements and the reachability labels. The calibration and test set contains respectively 8.5K and 10K samples.
    \item \textit{Sequential: } for the train set, 5K states are randomly sampled. From each of these states we simulate a long trajectory. From each long trajectory we obtain 100 sub-trajectories of length $32$ in a sliding window fashion. The same procedure is applied to the test and calibration set, where the number of initial states is respectively 1K and 850.
\end{itemize}
Data are scaled to the interval $[-1,1]$ to avoid sensitivity to different scales.
\review{While the chosen datasets are not too large, our approach would work well even with smaller datasets, resulting however in lower accuracy and higher uncertainty. 
In these cases, our proposed uncertainty-based active learning would represent the go-to solution as is designed for situations where data collection is particularly expensive.}



\paragraph{Computational costs.} 
NPM is designed to work at runtime in safety-critical applications, which translates in the need of high computational efficiency together with high reliability. The time needed to generate the dataset and to train both methods does not affect the runtime efficiency of the NPM, as it is performed only once (offline). 
Once trained, the time needed to analyse the reachability of the current sequence of observations is the time needed to evaluate one (or two) CNN, which is almost negligible (in the order of microseconds on GPU). 
On the other hand, the time needed to quantify the uncertainty depends on the size of the calibration set. 
This is one of the reasons that make active learning a preferable option, as it adds only the most significant points to the dataset. 
It is important to notice that the percentage of points rejected, meaning points with predictions estimated to be unreliable, affects considerably the runtime efficiency of the methods. Therefore, we seek a trade-off between accuracy and runtime efficiency.  Training the end-to-end approach takes around 15 minutes. Training the two-step approach takes around 40 minutes: 9 for the NSE, 11 for the NSC and 20 minutes for the fine-tuning. Making a single prediction takes around $7\times 10^{-7}$ seconds in the end-to-end scenario and $9\times 10^{-7}$ seconds in the two-step scenario. Training the SVC takes from 0.5 to 10 seconds, whereas computing values of confidence and credibility for a single point takes from 0.3 to 2 ms. Actively query new data from a pool of 50K samples takes around 5 minutes.

\paragraph{Performance measures.} The measures used to quantify the overall performance of the NPM under PO (both end-to-end and two-step)
are: the \emph{accuracy} of the reachability predictor, the \emph{error detection rate} and the \emph{rejection rate}. We seek high accuracies and detection rates without being overly conservative, meaning keeping a rejection rate as low as possible. We also check if and when the statistical guarantees are met empirically, via values of coverage and efficiency.
We analyse and compare the performances of NPM under PO on different configurations: an initial and active configuration for independent states and a temporally correlated (sequential) configuration. Additionally, we test the method for anomaly detection. 


\subsection{Results}
\paragraph{Initial setting.}
Table~\ref{table:orig_acc} compares the performances of the two approaches to PO-NPM via  predictive accuracy, detection rate, i.e. the percentage of prediction errors, either false-positives (FP) or false-negatives (FN), recognized by the error detection rule, and the overall rejection over the test set. We can observe how both methods work well despite PO, i.e., they reach extremely high accuracies and high detection rate. However, the two-step approach seems to behave slightly better than the end-to-end. As a matter of fact, accuracy is almost always greater than $99\%$ with a detection rate close to $100.00$. The average rejection rate is around $11\%$ in the end-to-end scenario, and reduces to $9\%$ in the two-step scenario, making the latter  less conservative ant thus more efficient from a computational point of view. 
These results come with no surprise, because, compared to the end-to-end one, the two-step approach leverages more information  available in the dataset for training, that is the exact sequence of states. 

\begin{table}
\centering
\begin{tabular}{ |c|r|r|r|r|r||r|r|r|r|r| }
\hline
&\multicolumn{5}{|c|}{End-to-end} & \multicolumn{5}{|c|}{Two-step}\\
\hline
\textbf{Model} & \textbf{Acc.} & \textbf{Det.} & \textbf{FN}&\textbf{FP}& \textbf{Rej.} &\textbf{Acc.} & \textbf{Det.} & \textbf{FN}&\textbf{FP} & \textbf{Rej.} \\
\hline
\textbf{SN} & 97.72 & 94.30 & 79/88 & 136/140 & 11.30 & 
97.12 & 95.49 & 53/54 & 222/234 & 19.98 \\
\textbf{IP} & 96.27 & 93.48 & 148/155  & 153/167&27.32 &
98.42 & 91.14 & 81/91 & 63/66 & 10.01\\

\textbf{CVDP} & 99.19 & 100.00 &30/30 & 51/51 & 5.75 &
99.68 & 100.00 &17/17 & 15/15 & 3.51  \\
\textbf{TWT} & 98.93 & 95.51 & 18/20 & 67/69 & 7.45 &
98.93 & 96.26 & 52/56 & 51/51 & 10.46 \\
\textbf{LALO} & 98.88 & 99.11 & 66/66 & 45/46 & 7.39 &
99.24 & 100.00 & 52/52  & 24/24 & 6.11 \\
\textbf{HC} & 99.63 & 100.00 & 19/19 & 15/15 & 8.47 &
99.84 & 100.00 & 8/8 & 8/8 & 4.03 \\
\hline
\end{tabular}
\caption{\textbf{Initial results}: \textit{Acc.} is the accuracy of the PO-NPM, \textit{Det.} the detection rate, \textit{Rej.} the rejection rate of the error detection rule and \textit{FN} (\textit{FP}) is the number of detected false negative (positive) errors.\vspace{-0.5cm}} \label{table:orig_acc}
\end{table}

\paragraph{Benefits of active learning.}
Table~\ref{table:active_acc} presents the results after one iteration of active learning. Additional data were selected from a pool of 50K points, using the error detection rule as query strategy. We observe a slight improvement in the performance, mainly reflected in higher detection rates and smaller rejection rates, with an average that reduces to $8\%$ for the end-to-end and to $6\%$ for the two-step. 

\begin{table}
\centering
\begin{tabular}{ |c|r|r|r|r|r||r|r|r|r|r| }
\hline
&\multicolumn{5}{|c|}{End-to-end} & \multicolumn{5}{|c|}{Two-step}\\
\hline
\textbf{Model} & \textbf{Acc.} & \textbf{Det.} & \textbf{FN}&\textbf{FP}& \textbf{Rej.} &\textbf{Acc.} & \textbf{Det.} & \textbf{FN}&\textbf{FP} & \textbf{Rej.} \\
\hline
\textbf{SN} & 98.06 & 94.87 & 81/88 & 104/107 & 9.80
& 98.41 & 100.00 & 55/55 & 104/104 & 12.00 \\
\textbf{IP} & 99.47 & 87.91 & 150/166 & 119/140 & 15.44
& 98.75 & 92.86 & 63/69 & 52/56 & 7.72\\
\textbf{CVDP} & 99.10 & 95.55 & 43/46 & 43/44 & 4.81
& 99.69 & 100.00 & 19/19 & 12/12 & 2.48 \\
\textbf{TWT} & 99.04 & 100.00 & 45/45 & 62/62 & 10.45
& 99.07 & 94.62 & 44/49 & 44/44 & 6.20\\
\textbf{LALO}& 98.79 & 96.69 &87/90 & 30/31 & 6.88  
& 99.27 &100.00& 40/40 & 33/33 & 4.28 \\
\textbf{HC}& 99.86 & 100.00 & 5/5 & 9/9 & 2.35
& 99.79 & 100.00 & 17/17 & 4/4 & 2.73 \\
\hline
\end{tabular}
\caption{\textbf{Active results} (1 iteration): \textit{Acc.} is the accuracy of the PO-NPM, \textit{Det.} the detection rate, \textit{Rej.} the rejection rate of the error detection rule and \textit{FN} (\textit{FP}) is the number of detected false negative (positive) errors. \vspace{-0.5cm}} \label{table:active_acc}
\end{table}

\paragraph{Probabilisic guarantees.}
In our experiments, we measured the efficiency as the percentage of singleton prediction regions over the test set.
Table~\ref{table:ponsc_stat_guar} compares the empirical coverage and the efficiency of the CP prediction regions in the initial and active scenario for both the end-to-end and two-step classifiers. The confidence level is set to $(1-\epsilon) = 95\%$. Fig.~\ref{fig:varying_eps} in Appendix~\ref{sec:var_eps} shows coverage and efficiency for different significance levels (ranging from 0.01 to 0.1). CP provides theoretical guarantees on the validity, meaning empirical coverage matching the expected one of $95\%$, only in the initial setting. As a matter of fact, with active learning we modify the data-generating distribution of the training and calibration sets,  while the test set remains the same, i.e., sampled from the original data distribution. As a result, we observe (Table~\ref{table:ponsc_stat_guar}) that both methods in the initial setting are valid. In the active scenario, even if theoretical guarantees are lost, we obtain both better coverage and higher efficiency. This means that the increased coverage is not due to a more conservative predictor but to an improved accuracy. 

\begin{table}
\centering
\begin{tabular}{ |c|r|r|r|r||r|r|r|r|}
\hline
&\multicolumn{4}{|c|}{End-to-end}&\multicolumn{4}{|c|}{Two-step} \\
\hline
& \multicolumn{2}{|c|}{initial} & \multicolumn{2}{|c|}{active} & \multicolumn{2}{|c|}{initial} & \multicolumn{2}{|c|}{active}  \\
\hline
\textbf{Model}  & \textbf{Cov.} & \textbf{Eff.}& \textbf{Cov.} & \textbf{Eff.}& \textbf{Cov.} & \textbf{Eff.}& \textbf{Cov.} & \textbf{Eff.}\\
\hline
\textbf{SN} & 95.12 & 95.70 & 97.19 & 98.50 & 94.80 & 99.54 & 97.32 & 98.37 \\
\textbf{IP} & 95.30& 89.31 & 96.60 & 99.62 & 94.85 & 94.92 & 97.28 &97.88 \\
\textbf{CVDP} & 95.73 & 95.73 & 98.00 & 98.02 & 95.63 & 95.63 & 98.31 & 98.34 \\
\textbf{TWT} & 96.43 & 96.43 & 99.99 & 97.26 & 96.60 & 96.97 & 99.66 & 97.20\\
\textbf{LALO} & 94.59 & 94.61 & 97.28 & 98.52 & 94.66 & 94.66 & 97.48 & 97.55\\
\textbf{HC} & 95.03 & 95.03 & 97.65 & 97.65 & 94.97 & 94.97 & 97.69 & 97.69 \\
\hline
\end{tabular}
\caption{Coverage and efficiency for both the approaches to PO-NPM. Initial results are compared with results after one active learning iteration. Expected coverage $95\%$.\vspace{-0.5cm}}\label{table:ponsc_stat_guar}
\end{table}

Table~\ref{table:nsc_nse_stat_guar} shows values of coverage and efficiency for the two separate steps (state estimation and reachability prediction) of the two-step approach. Recall that the efficiency in the case of regression, and thus of state estimation, is given by the volume of the prediction region. So, the smaller the volume, the more efficient the regressor. 
The opposite holds for classifiers, where a large value of efficiency means tight prediction regions. It is interesting to observe how active learning makes the NSC reach higher coverages at the cost of more conservative prediction regions (lower efficiency), 
whereas the NSE coverage is largely unaffected by  active learning (except for TWT). Reduction in NSC efficiency, differently from the two-step combined approach, is likely due to an adaptation of the method to deal with and correct noisy estimates. 
Such behaviour suggests that the difficulty in predicting the reachability of a certain state is independent of how hard it is to reconstruct that state\footnote{We select re-training points based on the uncertainty of the reachability predictor; if the SE performed badly on those same points, re-training would have led to a higher SE accuracy and hence, increased coverage.}.

\begin{table}
\centering
\begin{tabular}{ |c|r|r|r|r||r|r|r|r|}
\hline
&\multicolumn{4}{|c|}{NSC}&\multicolumn{4}{|c|}{NSE} \\
\hline
& \multicolumn{2}{|c|}{initial} & \multicolumn{2}{|c|}{active} & \multicolumn{2}{|c|}{initial} & \multicolumn{2}{|c|}{active}  \\
\hline
\textbf{Model}  & \textbf{Cov.} & \textbf{Eff.}& \textbf{Cov.} & \textbf{Eff.}& \textbf{Cov.} & \textbf{Eff.}& \textbf{Cov.} & \textbf{Eff.}\\
\hline
\textbf{SN} & 94.82 & 99.51 & 97.23 & 90.12 & 94.49 & 1.361 & 95.18 & 1.621 \\
\textbf{IP} & 94.51 & 99.69 & 97.23 & 91.63 & 94.65 & 3.064 & 95.44 & 3.233  \\
\textbf{CVDP} & 95.60 & 95.64 & 98.25 & 98.32 & 95.37 & 0.343 & 96.40 & 0.358  \\
\textbf{TWT} & 96.68 & 96.98 & 98.72 & 95.61 & 95.07 & 0.770 & 100.00 & 1.366 \\
\textbf{LALO} & 94.88 & 98.18 & 98.01 & 80.86 & 95.29 & 0.6561 & 95.36 & 0.8582 \\
\textbf{HC} & 94.67 & 94.74 & 97.33 & 99.12 & 94.50 & 12.44 & 94.58  & 12.464  \\
\hline
\end{tabular}
\caption{Coverage and efficiency for the two steps of the two-step approach. NSC is a classifier, whereas NSE is a regressor. Initial results are compared with results after one active learning iteration. Expected coverage $95\%$.\vspace{-0.5cm}}\label{table:nsc_nse_stat_guar}
\end{table}

\begin{figure}[ht]
    \centering
    \includegraphics[scale=0.25]{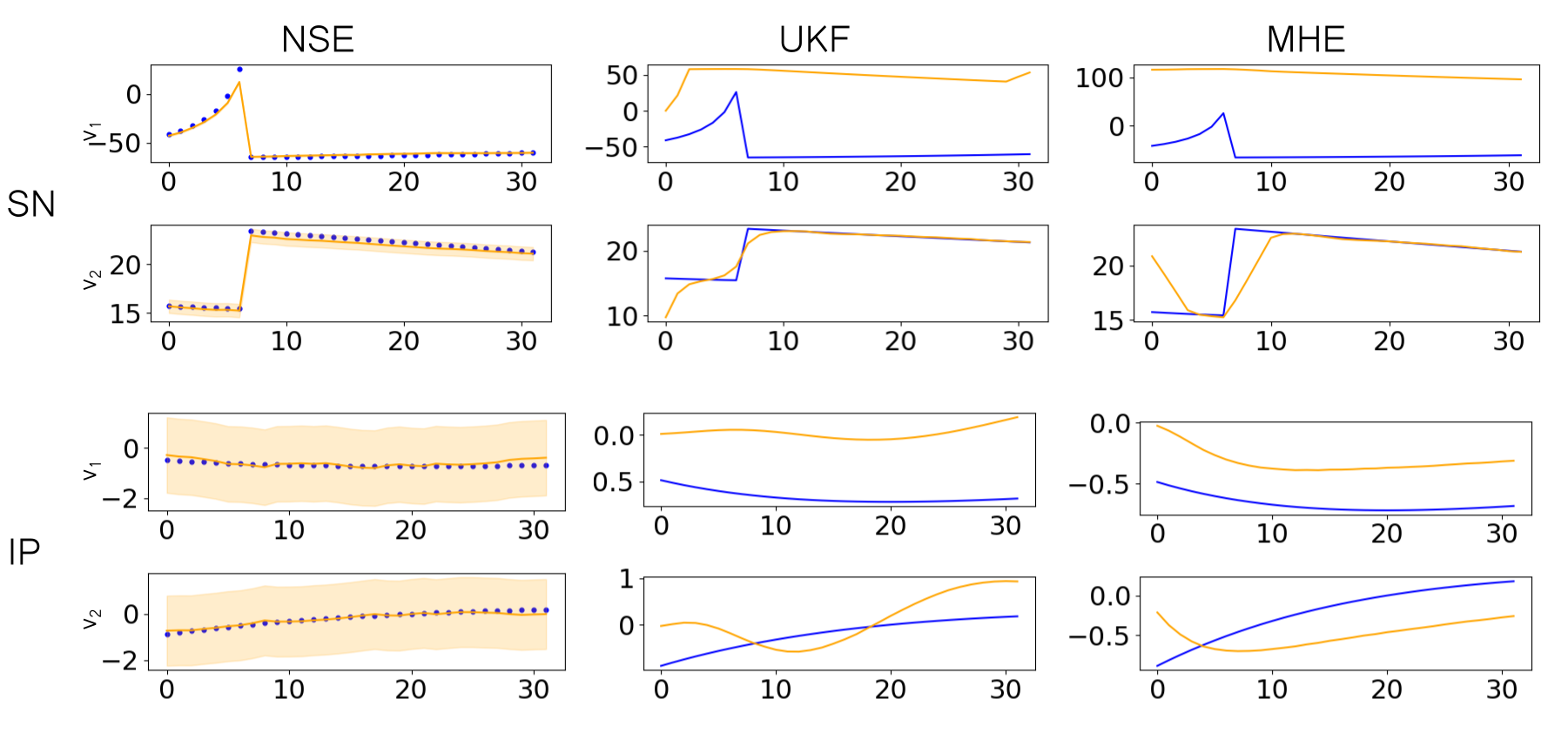}
    \caption{Comparison of different state estimators on a state of the SN (top) and IP (bottom) model. Blue is the exact state sequence, orange is the estimated one. 
    \vspace{-0.5cm}}\label{fig:sn_se}
\end{figure}

\paragraph{State estimator.} We compare the performances of the NSE with two traditional state estimation techniques: Unscented Kalman Filters\footnote{pykalman library: https://pykalman.github.io/} (UKF)~\cite{wan2000unscented} and Moving Horizon Estimation\footnote{do-mpc library: https://www.do-mpc.com/en/latest/} (MHE)~\cite{allan2019moving}. In particular, for each point in the test set we compute the relative error given by the norm of the difference between the real and reconstructed state trajectories divided by the maximum range of state values. 
The results, presented in full in Appendix~\ref{sec:SE}, show how our neural network-based state estimator significantly outperforms both UKF and MHE in our case studies. 
Moreover, unlike the existing SE approaches, our state estimates come with a prediction region that provides probabilistic guarantees on the expected reconstruction error, as shown in Fig.~\ref{fig:sn_se}. 

\paragraph{Sequential data.} All the results presented so far consider a dataset $\mathcal{D}_{PO-NPM}$ of observation sequences generated by independently sampled initial states. However, we are interested in applying NPM at runtime to systems that are evolving in time. States will thus have a temporal correlation, meaning that we lose the exchangeability requirement behind the theoretical validity of CP regions. 
Table~\ref{table:seq_res} shows the performance of predictor and error detection 
trained and tested on sequential data. 
In general, accuracy and detection rates are still very high (typically above $95\%$), but the results are on average worse than the independent counterpart. The motivation could be two-fold: on one side, it is reasonable to assume that a recurrent neural net would perform better on sequential data, compared to CNN, on the other, the samples contained in the sequential dataset are strongly correlated and thus they may cover only poorly the state space. 
The table also shows values of coverage and efficiency of both the end-to-end and the two-step approach. Even if theoretical validity is lost, we still observe empirical coverages that match the nominal value of $95\%$, i.e., the probabilistic guarantees are satisfied in practice. 

\begin{table}
\centering
\begin{tabular}{ |c|r|r|r|r|r||r|r|r|r|r| }
\hline
&\multicolumn{5}{|c|}{End-to-end} & \multicolumn{5}{|c|}{Two-step}\\
\hline
\textbf{Model} & \textbf{Acc.} & \textbf{Det.} &  \textbf{Rej.} & \textbf{Cov.}&\textbf{Eff.}& \textbf{Acc.} & \textbf{Det.} & \textbf{Rej.} & \textbf{Cov}&\textbf{Eff.} \\
\hline
\textbf{SN} & 94.96 & 85.83 & 19.74 & 93.93 &97.73 & 90.37 & 81.93 &26.59 &95.01  & 88.66\\ 
\textbf{IP} & 94.17 & 91.08 &  31.74 &95.31 & 84.32 & 91.47 &98.01 & 30.81 &95.23  & 90.23\\ 
\textbf{CVDP} &98.97 & 99.12 & 7.97& 94.88 &94.92 & 98.33 & 98.20 & 9.89 & 94.89 & 95.19   \\ 
\textbf{TWT} & 96.95 &95.33 & 16.84 & 93.42 & 94.52 & 95.74 & 92.72 & 23.52 & 93.60 & 96.16   \\ 
\textbf{LALO} & 98.99 &97.75 &7.18 & 95.93 & 97.08 & 99.26 & 100.00& 5.37& 95.78&95.80  \\ 
\textbf{HC} & 99.57 & 100.00 & 3.89 & 94.29 & 94.29 & 99.64 & 97.22 & 3.84 & 94.51 & 94.52   \\ 
\hline
\end{tabular}
\caption{\textbf{Sequential results}: \textit{Acc.} is the accuracy of the PO-NPM, \textit{Det.} the detection rate, \textit{Rej.} the rejection rate, \textit{Cov.} the CP coverage and \textit{Eff.} the CP efficiency. \vspace{-0.5cm}}\label{table:seq_res}

\end{table}

\begin{figure}
    \centering
    \includegraphics[scale=0.22]{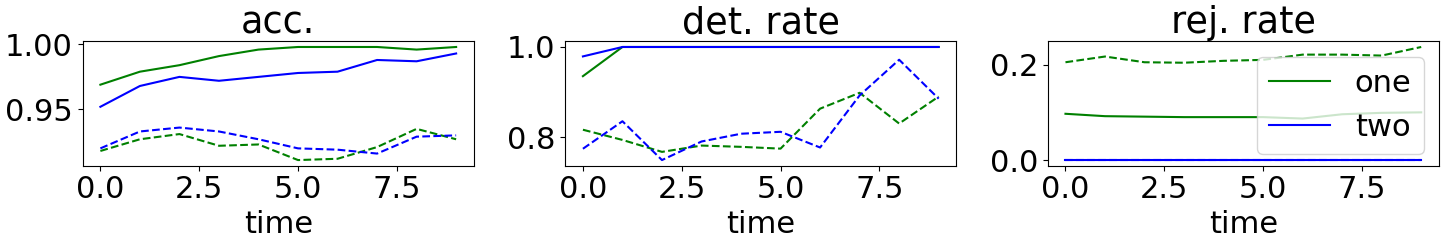}
    \caption{\textbf{Anomaly detection} (TWT model). Dashed lines denotes the performances on observations with anomaly in the noise. Blue is for the two-step approach, green for the end-to-end. \vspace{-0.5cm}}
    \label{fig:anomalies}
\end{figure}

\paragraph{Anomaly detection.} The data-generating distribution at runtime is assumed to coincide with the one used to generate the datasets. However, in practice, such distribution is typically unknown and subject to runtime deviations. Thus, we are interested to observe how the sequential PO-NPM behave when an anomaly takes place. In our experiments, we model an anomaly as an increase in the variance of the measurement noise, i.e. $\mathcal{W}' = \mathcal{N}(0,0.25\cdot I)$. Fig.~\ref{fig:anomalies} compares the performances with VS without anomaly on a single case study (the other case studies are shown in Appendix~\ref{sec:anomaly_plots}). We observe that the anomaly causes a drop in accuracy and error detection rate, which comes with an increase in the number of predictions rejected because deemed to be unreliable. These preliminary results show how an increase in the NPM rejection rate could be used as a significant measure to preemptively detect runtime anomalies.

\section{Related work} 
Our approach extends and generalize neural predictive monitoring~\cite{bortolussi2019neural,bortolussineural} to work under partial observability. To our knowledge, the only existing work to focus on PM and PO is~\cite{chou2020predictive}, which combines Bayesian estimation with pre-computed reach sets to reduce the runtime overhead. While their reachability bounds are certified, no correctness guarantees can be established for the estimation step. 
Our work instead provides probabilistic guarantees as well as techniques for preemptive error detection. A related but substantially different problem is to verify signals with observation gaps using state estimation to fill the gaps~\cite{stoller2011runtime,kalajdzic2013runtime}. 

\review{In~\cite{pinisetty2017predictive} a model-based approach to predictive runtime verification is presented. However, PO and computational efficiency are not taken into account. A problem very similar to ours is addressed in~\cite{junges2021runtime}, but for a different class of systems (MDPs). 
}

Learning-based approaches for reachability prediction of hybrid and stochastic systems include~\cite{bortolussi2016smoothed,phan2018neural,djeridane2006neural,royo2018classification,yel2020assured,granig2020weakness}. Of these,~\cite{yel2020assured} develop, akin to our work, error detection techniques, but using neural network verification methods~\cite{ivanov2019verisig}. Such verification methods, however, do not scale well on large models and support only specific classes of neural networks. On the opposite, our uncertainty-based error detection can be applied to any ML-based predictive monitor. 
Learning-based PM approaches for temporal logic properties~\cite{qin2019predictive,DBLP:journals/corr/abs-2011-00384} typically learn a time-series model from past observations and then use such model to infer property satisfaction. In particular,~\cite{qin2019predictive} provide (like we do) guaranteed prediction intervals, but (unlike our method) they are limited to ARMA/ARIMA models. Ma et al~\cite{DBLP:journals/corr/abs-2011-00384} use uncertainty quantification with Bayesian RNNs to provide confidence guarantees. However, these models are, by nature, not well-calibrated (i.e., the model uncertainty does not reflect the observed one~\cite{kuleshov2018accurate}), making the resulting guarantees not theoretically valid\footnote{The authors develop a solution for Bayesian RNNs calibration, but such solution in turn is not guaranteed to produce well-calibrated models.}.

PM is at the core of the Simplex architecture~\cite{sha2001using,johnson2016real} and recent extensions thereof~\cite{phan2020neural,mehmood2020distributed}, where the PM component determines when to switch to the fail-safe controller to prevent imminent safety violations. In this context, our approach can be used to guarantee arbitrarily small probability of wrongly failing to switch. 


\section{Conclusion}\label{sec:conclusion}

We presented an extension of the Neural Predictive Monitoring~\cite{bortolussineural} framework to work under the most realistic settings of noise and partially observability. We proposed two alternative solution strategies: an end-to-end solution, predicting reachability directly from raw observations, and a two-step solution, with an intermediate  state estimation step. Both methods produce extremely accurate predictions, with the two-step approach performing better overall than the end-to-end version, and further providing accurate reconstructions of the true state. The online computational cost is negligible, making this method suitable for runtime applications. 
The method is equipped with an error detection rule to prevent reachability prediction errors, as well as with prediction regions providing probabilistic guarantees.
We demonstrated that error detection can be meaningfully used for active learning, thereby improving our models on the most uncertain inputs. 

As future work, we plan to extend this approach to fully stochastic models, investigating the use of deep generative models for state estimation. 
We will further explore the use of recurrent or attention-based architectures in place of convolutional ones to improve  performance for sequential data.


\bibliographystyle{splncs04}
\bibliography{biblio}

\newpage

\appendix

\section{Models and Case Studies}\label{appendix:models}
We briefly introduce the case studies used in our experimental evaluation.

\subsubsection{Spiking Neuron} 
We consider the spiking neuron model on the Flow* website\footnote{\url{https://flowstar.org/examples/}{https://flowstar.org/examples/}}, describing the evolution of a neuron’s action potential. It is a hybrid system with one mode and one jump. The dynamics is defined by the ODE

\begin{equation}
\begin{cases}
\dot{s_2} &= 0.04s_2^2 + 5s_2 + 140 - s_1 + I \\
\dot{s_1} &= a \cdot \left(b \cdot s_2 - s_1\right)
\end{cases}
\end{equation}

The jump condition is $s_2 \geq 30$, and the associated reset is $s_2':= c \land s_1' := s_1 + d$, where, for any variable $x$, $x'$ denotes the value of $x$ after the reset.

The parameters are $a = 0.02$, $b = 0.2$, $c = -65$, $d = 8$, and $I = 40$ as reported on the Flow* website. We consider the unsafe state set $U = \left\lbrace \left( s_2, s_1 \right) \mid s_2 \leq 68.5 \right\rbrace$. This corresponds to a safety property that can be understood as the neuron does not undershoot its resting-potential region of $\lbrack -68.5, -60 \rbrack$.  
The domain for sampling is  $68.5 < s_2 \leq 30 \land 0 \leq s_1 \leq 25$.  We consider the unsafe set $Y$ defined by $v_2 \le 68.5$, expressing that the neuron should not undershoot its resting potential. The time bound for the reachability property is $H_f = 16$.  Given a state $s=(s_1,s_2)$ we observe a noisy measure of $s_2$, $y= s_2+w$, with $w\sim\mathcal{N}(0,0.1)$,  $H_p = 4$.

\subsubsection{Inverted Pendulum} We consider the classic inverted pendulum on a cart nonlinear system. This is a classic, widely used example of a non-linear system.  The control input $F$ is a force applied to the cart with the goal of keeping the pendulum in upright position, i.e., $\theta = 0$. The dynamics is given by
\begin{equation}
	J \cdot \ddot{\theta} = m \cdot l \cdot g \cdot \sin(\theta) - m \cdot l \cos(\theta) \cdot F
    \label{eq:ip}
\end{equation}
where $J$ is the moment of inertia, $m$ is the mass of the
pendulum, $l$ is the length of the rod, and $g$ is the gravitational acceleration.
We set $J = 1$, $m = 1/g$, $l = 1$, and let $u = F/g$. Eq.~\ref{eq:ip} becomes

\begin{equation}
\begin{cases}
    \dot{\theta} = \omega\\
    \dot{\omega} = \sin(\theta) -\cos(\theta) \cdot u
\end{cases}
\label{eq:ip-simplified}
\end{equation}

We consider the control law of Eq.~\ref{eq:ip-control-law}. We consider the unsafe state set $U = \left\lbrace \left( \theta, \omega \right) \mid \theta < -\pi/4 \lor \theta > \pi/4 \right\rbrace$. This unsafe region corresponds to the safety property that keeps the pendulum within $45^\circ$ of the vertical axis. The domain for sampling is  $\theta \in \lbrack -\pi/4, \pi/4 \rbrack \land \omega \in \lbrack -1.5, 1.5 \rbrack$.  We used time bound $H_f = 5$ and $H_p = 1$. 

\begin{equation}
u = 
\begin{cases}
    \displaystyle{\frac{2 \cdot \omega + \theta + \sin(\theta)}{\cos(\theta)}}, & E \in \lbrack -1, 1 \rbrack, \mid \omega \mid + \mid \theta \mid \leq 1.85 \\[10pt]
    0, & E \in \lbrack -1, 1 \rbrack, \mid \omega \mid + \mid \theta \mid > 1.85 \\[10pt]
    \displaystyle{\frac{\omega}{1 + \mid \omega \mid}  \cos(\theta)}, & E < -1 \\[10pt]
    \displaystyle{\frac{-\omega}{1 + \mid \omega \mid}  \cos(\theta)}, & E > 1 \\
\end{cases}
\label{eq:ip-control-law}
\end{equation}
where $E = 0.5 \cdot \omega + (\cos(\theta) - 1)$ is the pendulum energy.

We consider the unsafe set $U$ defined by $|\theta| > \pi/6$, corresponding to the safety
property that keeps the pendulum within $30^\circ$ of the vertical axis. The time bound is $H_f = 5$. 
Given a state $s=(s_1,s_2)$, we observe a noisy measure of the energy of the system $y = s_2/2+\cos(s_1)-1 + w$, where $w\sim\mathcal{N}(0,0.005)$.

\subsubsection{Laub-Loomis} This model (from ARCH-COMP20~\cite{ernst2020arch}) studies a class of enzymatic activities. The dynamics can be defined by the following ODE:
\begin{equation}
\begin{cases}
\dot{s}_1 &= 1.4s_3-0.9s_1\\
\dot{s}_2 &= 2.5s_5-1.5s_2\\
\dot{s}_3 &= 0.6s_7-0.8s_2s_3\\
\dot{s}_4 &= 2-1.3s_3s_4\\
\dot{s}_5 &= 0.7s_1-s_4s_5\\
\dot{s}_7 &= 0.3s_1-3.1s_6\\
\dot{s}_8 &= 1.8s_6-1.5s_2s_7.\\
\end{cases}
\end{equation}
The system is asymptotically stable with equilibrium at the origin.
The unsafe region is defined as $U = \{s: s_4\ge 4.5\}$.
Given a state $s = (s_1,\dots , s_7)$ we observe $y = (s_1, s_2, s_3, s_5, s_6, s_7)+w$, with $w\sim \mathcal{N}(0,0.01)$, $H_p = 5$ and $H_f = 20$.

\subsubsection{Coupled Van Der Pol} This benchmark (from ARCH-COMP20~\cite{ernst2020arch}) consists of two coupled oscillators. The dynamics can be defined by the following ODE:
\begin{equation}
\begin{cases}
\dot{s}_1 &= s_2\\
\dot{s}_2 &= (1-s_1^2)s_2-2s_1+s_3\\
\dot{s}_3 &= s_4\\
\dot{s}_4 &= (1-s_3^2)s_4-2s_3+s_1\\
\end{cases}
\end{equation}
Given a state $s=(s_1,s_2,s_3,s_4)$ we observe $y = (s_1,s_3)+ w$, with $w\sim\mathcal{N}(0,0.01)$. Unsafe region $ U = \{s: s_{2} \ge 2.75 \land s_{2} \ge 2.75\}$. $H_p = 8$, $H_f = 7$.

\subsubsection{Triple Water Tank}
In the TWT, three water tanks are connected by pipes, and the water level in each tank is separately controlled by the pump in the tank, which can be turned on or off.
The water level of each tank depends on the mode $q\in\{on,off\}$ of the tank and the levels of the adjacent tanks. The water level $v_i$ of tank $i$ changes according to the ordinary differential equations:
\begin{align}
    A_i\dot{v}_i &= m_i+a\sqrt{2gv_{i-1}}-b\sqrt{2gv_i}\quad &\mbox{ if } q_i = on\\
    A_i\dot{v}_i &= a\sqrt{2gv_{i-1}}-b\sqrt{2gv_i}\quad &\mbox{ if } q_i = off,\\
\end{align}
where $A_i,m_i,a,b$ are constants determined by the size of the tank, the power of the pump, the width of the I/O pipe, and g is the standard gravity constant. We set $v_0=0$ for the leftmost tank $1$.
For the TWT model\footnote{\url{http://dreal.github.io/benchmarks/networks/water/}}, $U$ is given by states where the water level of any of the tanks falls outside a given safe interval $I$, i.e., $U = \vee_{i=1}^3 x_i \not\in I$, where $x_i$ is the water level of tank $i$. The state distribution considers water levels uniformly distributed within the safe interval. The time bound is $H_f=H_p=1$.

\subsubsection{Helicopter Controller} We augment the 28-variable helicopter controller available on
SpaceEx website with a variable $z$ denoting the helicopter’s altitude. The dynamics of $z$ is given by $\dot{z} = v_z$ , where $v_z$ is the vertical velocity and represented by variable $x_8$. The unsafe set $D$ is defined by $z \le 0$. The time bound is $H_f = 5$. Since this model is
large and publicly available on SpaceEx website, we do not provide the details here.

Adam~\cite{bengio2015rmsprop} is the algorithm used to optimize every loss. In the end-to-end approach the learning rate is set to $10^{-5}$ and it is trained for 200 epochs with batches of size 64. In the two-step approach the  learning  rate  is  set  to  $10^{-6}$ when training NSC and NSE separately and to $10^{-7}$ for the combined fine-tuning phase. The NSE and the NSC are trained for 200 epochs on batches of size 64 and, finally, 100 epochs of fine-tuning are performed.

\section{Runtime Performances and Anomaly Detection}\label{sec:anomaly_plots}

\begin{figure}
    \centering
 \includegraphics[scale=0.3]{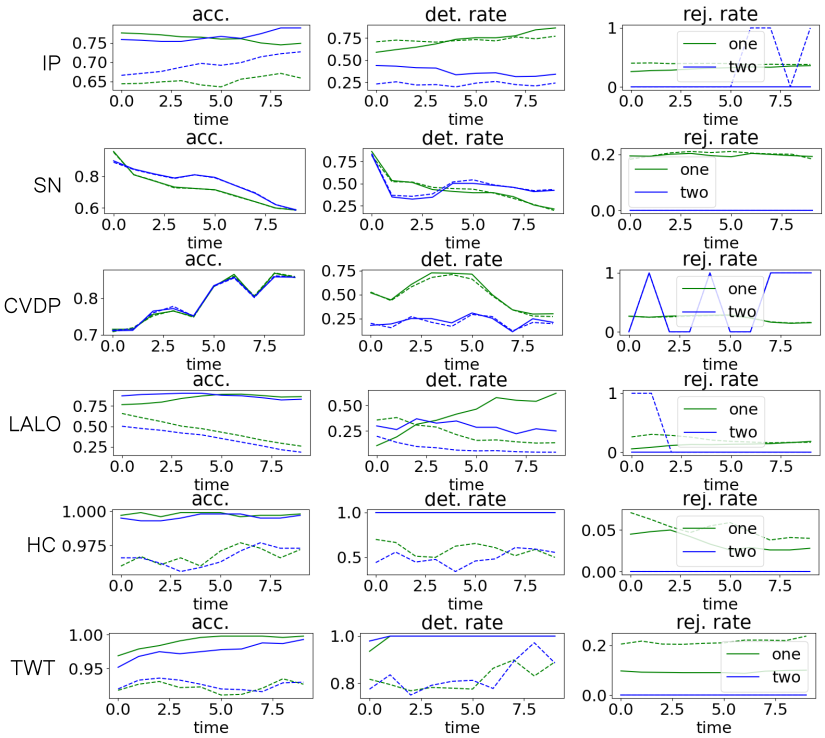}
    \caption{\textbf{Anomaly detection} (TWT model). Blue lines denotes the performances of the two-step approach. Green line the end-to-end approach. Dashed lines denotes the performances on observations with anomaly in the noise.}
    \label{fig:anomaly}
\end{figure}

\section{Coverage and Efficiency for varying $\epsilon$}\label{sec:var_eps}

\begin{figure}
    \centering
    \includegraphics[scale =0.26]{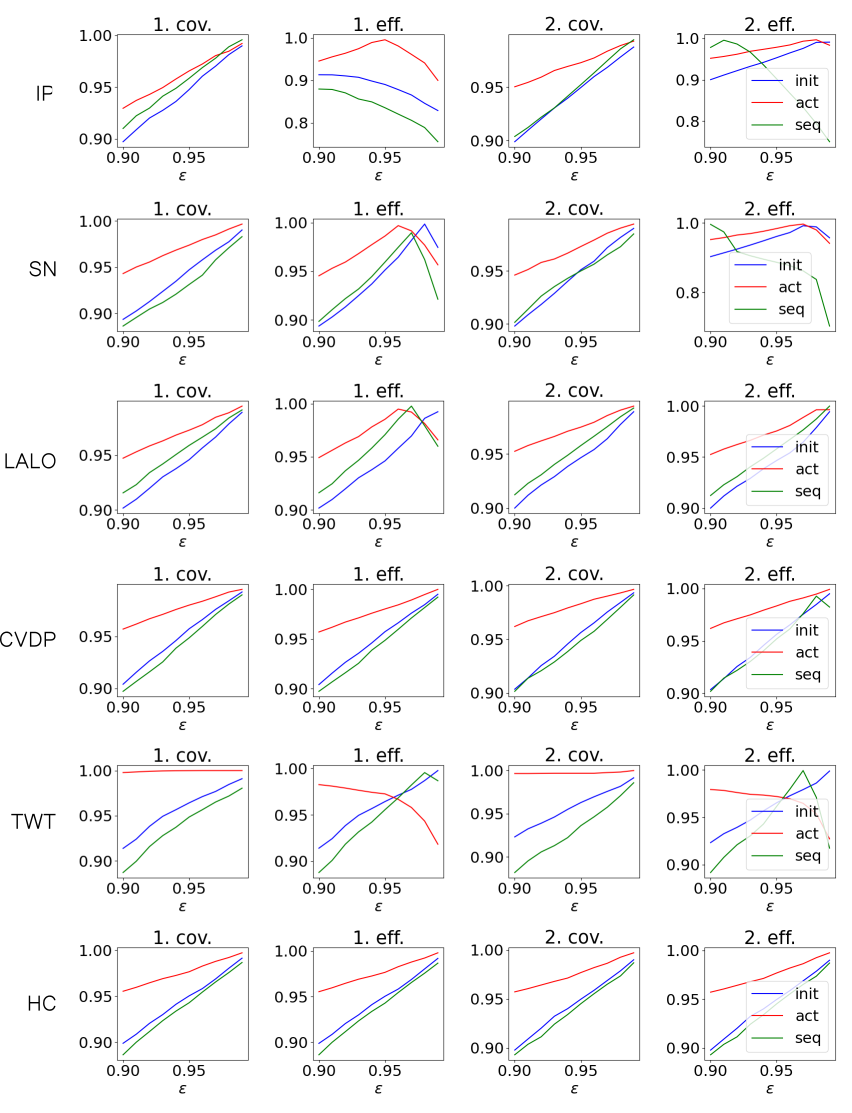}
    \caption{Coverage and efficiency of the PO-NSC for the initial, active and sequential configuration. 1. in the title denotes the end-to-end approach, whereas 2. denotes the two-step approach.}
    \label{fig:varying_eps}
\end{figure}

\section{Architecture and training details}\label{sec:arch}

Both approaches to PO-NPM consider sequences of states and observations of fixed length, thus one-dimensional CNNs are indeed a suitable architecture. In particular, the end-to-end classifier and the NSC share the same architecture: four convolutional layers with 128 filters of size 3, with Leaky-ReLU activation functions with slope 0.2 and, for regularizaion purposes, a drop-out with probability 0.2. The architectures terminates with two dense layers with 100 and 2 nodes respectively. The last layer has a ReLU activation function, to enforce positivity of the class likelihood scores. 

On the other hand, the NSE architecture is composed of 5 convolutional layers with 128 filters of size 5, LeakyReLU activations with slope 0.2 and drop-out with probability 0.2. The last layer has the Tanh as activation function, so that the reconstructed states is bounded to the interval $[-1,1]$.

\section{Comparison of State Estimators}\label{sec:SE}

\begin{figure}[ht]
    \centering
    \includegraphics[scale=0.23]{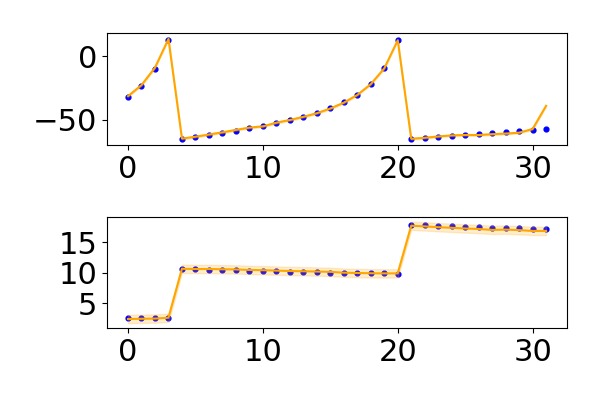}
    \includegraphics[scale=0.23]{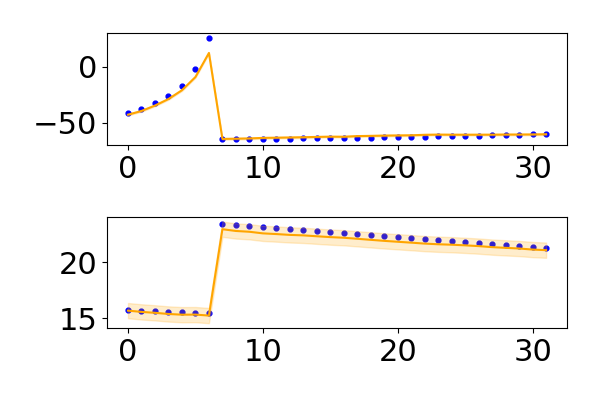}
    \includegraphics[scale=0.23]{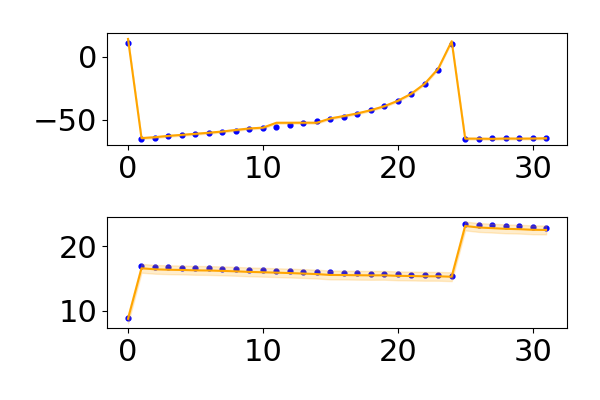}
    \caption{SN: \textbf{Neural SE}. Each column is a different test point and each row is a variable of the state space.}
    
    \includegraphics[scale=0.23]{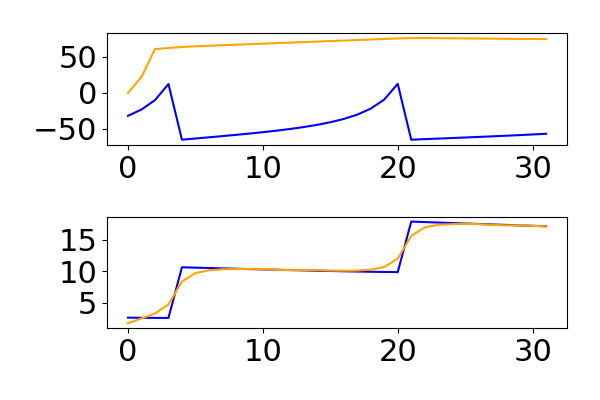}
    \includegraphics[scale=0.23]{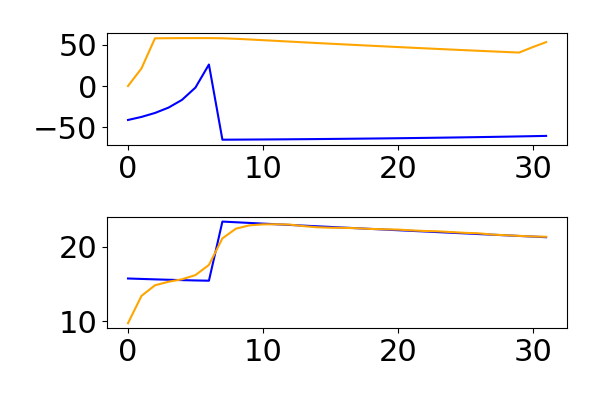}
    \includegraphics[scale=0.23]{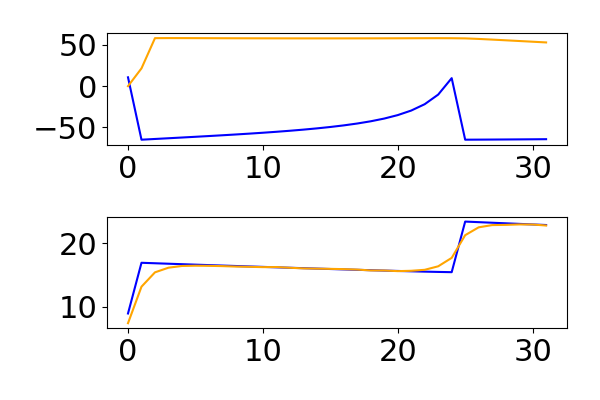}
    
    \caption{SN: \textbf{Unscented Kalman Filters}. Each column is a different test point and each row is a variable of the state space.}
    \includegraphics[scale=0.23]{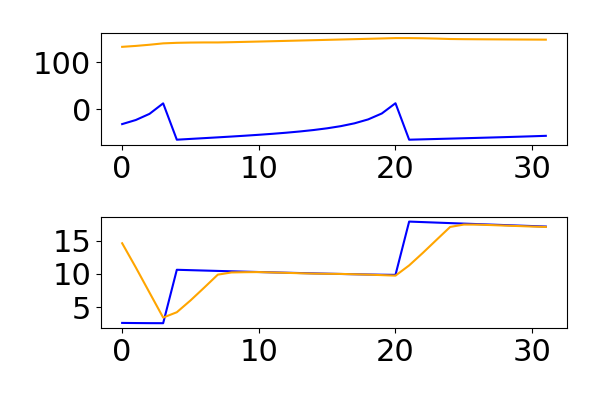}
    \includegraphics[scale=0.23]{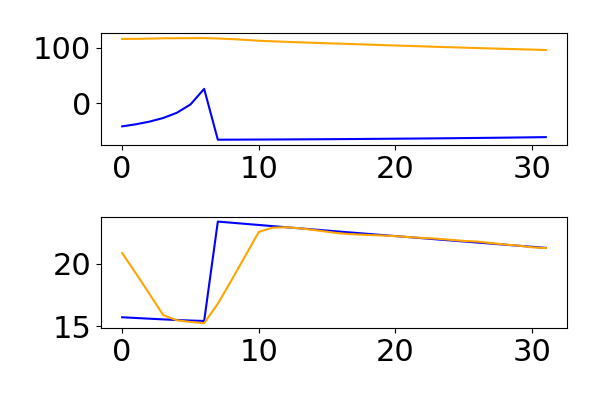}
    \includegraphics[scale=0.23]{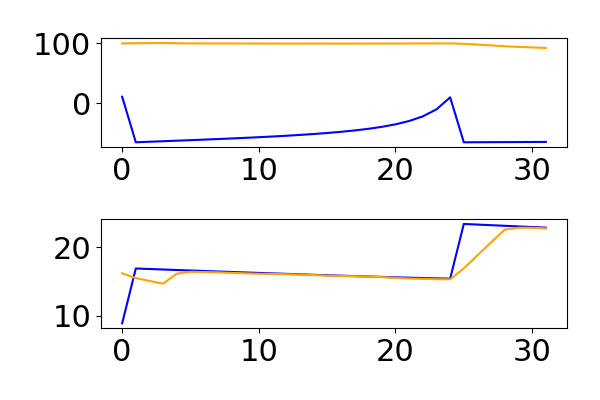}
    \caption{SN: \textbf{Moving Horizon Estimate}. Each column is a different test point and each row is a variable of the state space.}
\end{figure}

\begin{figure}[ht]
    \centering
    \includegraphics[scale=0.25]{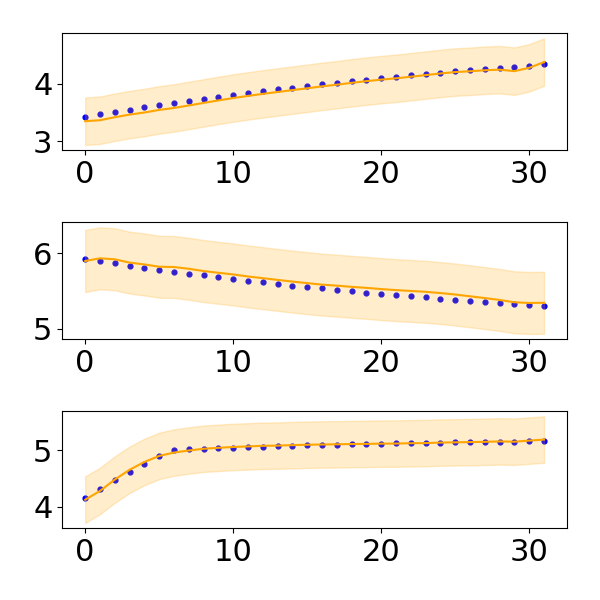}
    \includegraphics[scale=0.25]{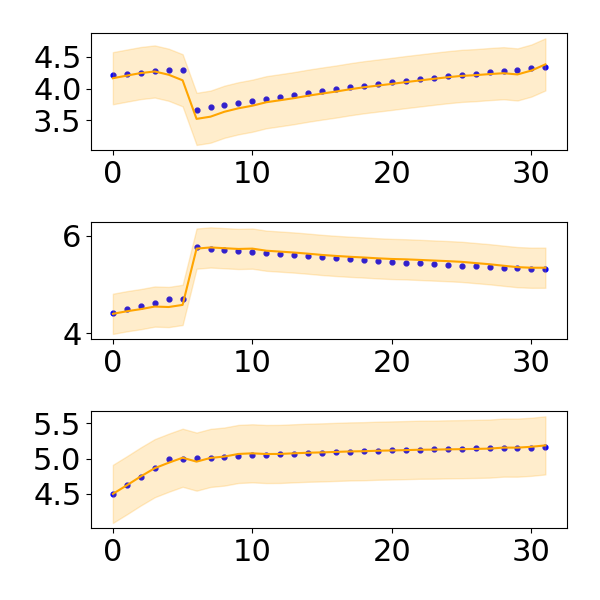}
    \includegraphics[scale=0.25]{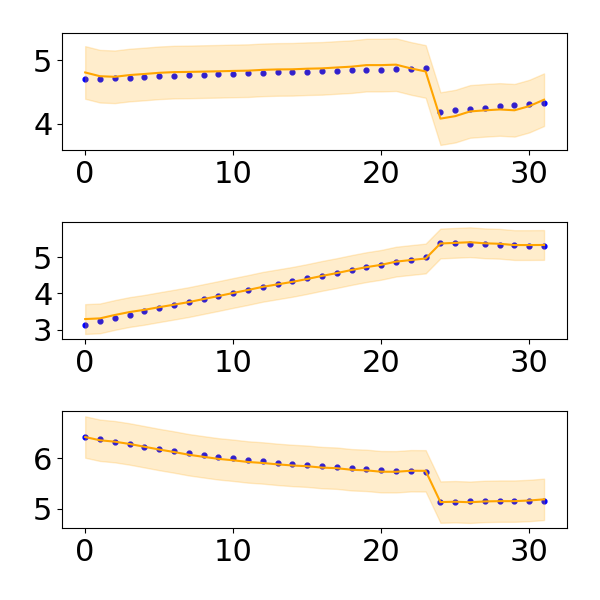}
    
    \caption{TWT: \textbf{Neural SE}. Each column is a different test point and each row is a variable of the state space.}
    \includegraphics[scale=0.25]{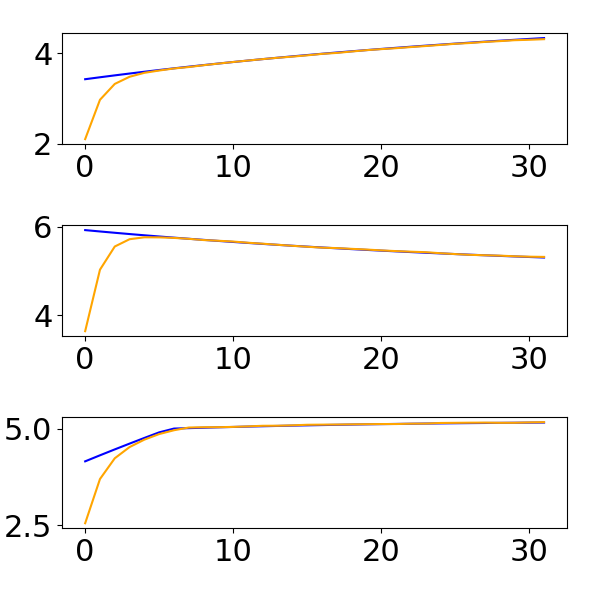}
    \includegraphics[scale=0.25]{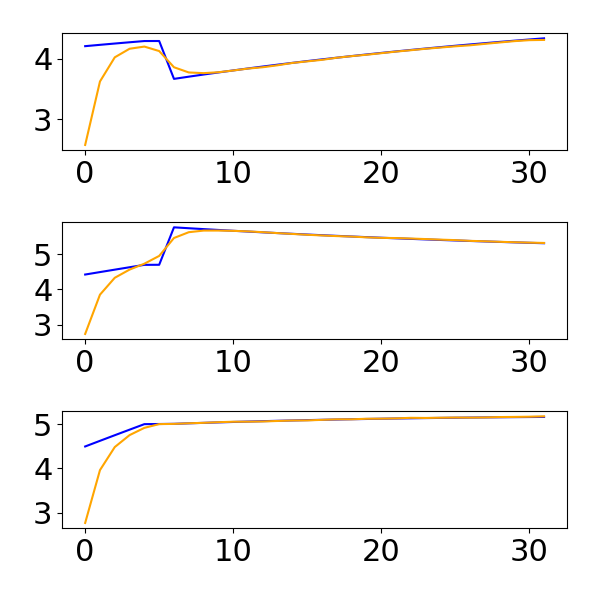}
    \includegraphics[scale=0.25]{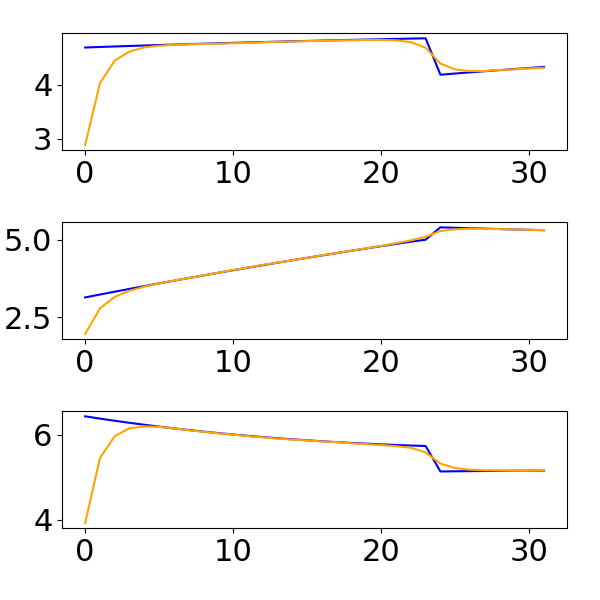}
    \caption{TWT: \textbf{Unscented Kalman Filters}. Each column is a different test point and each row is a variable of the state space.}

    \includegraphics[scale=0.25]{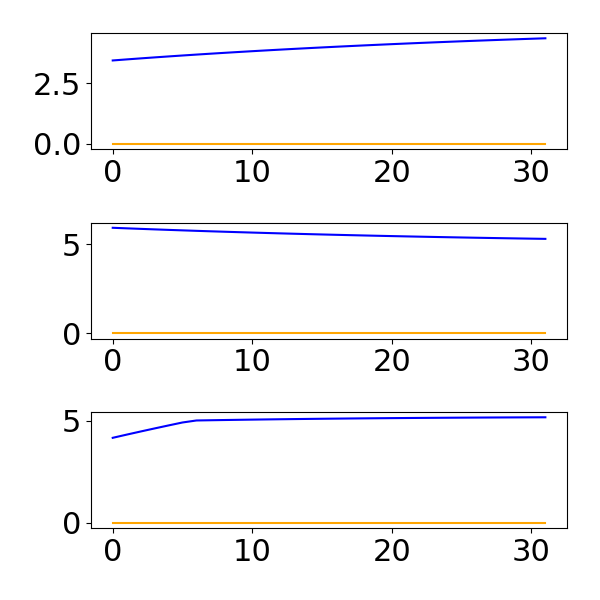}
    \includegraphics[scale=0.25]{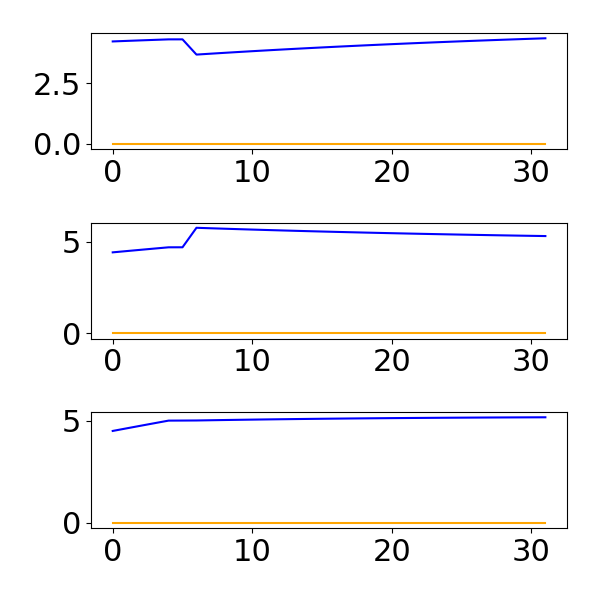}
    \includegraphics[scale=0.25]{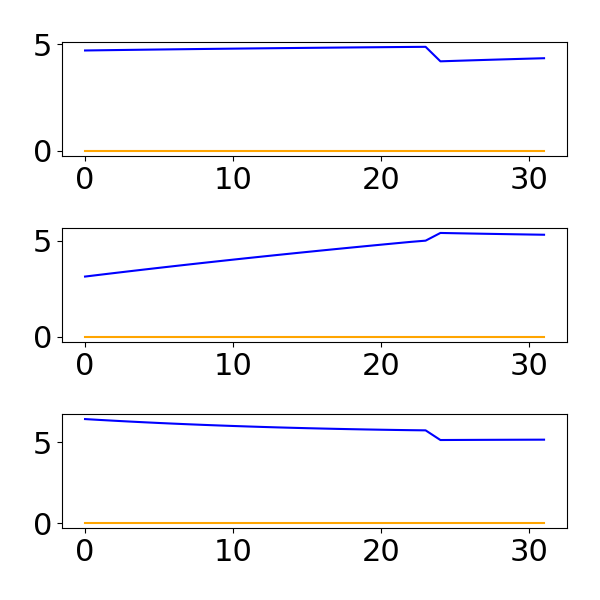}
    \caption{TWT: \textbf{Moving Horizon Estimate}. Each column is a different test point and each row is a variable of the state space.}

\end{figure}

\begin{figure}[ht]
    \centering
\includegraphics[scale=0.24]{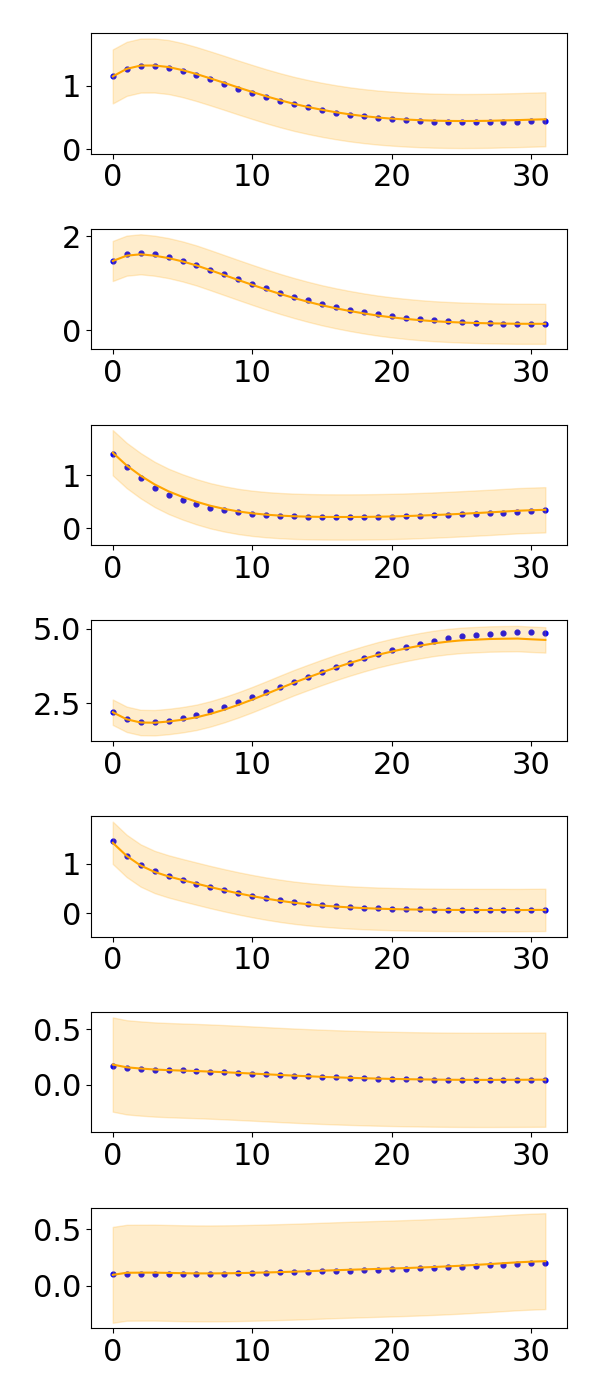}
    \includegraphics[scale=0.24]{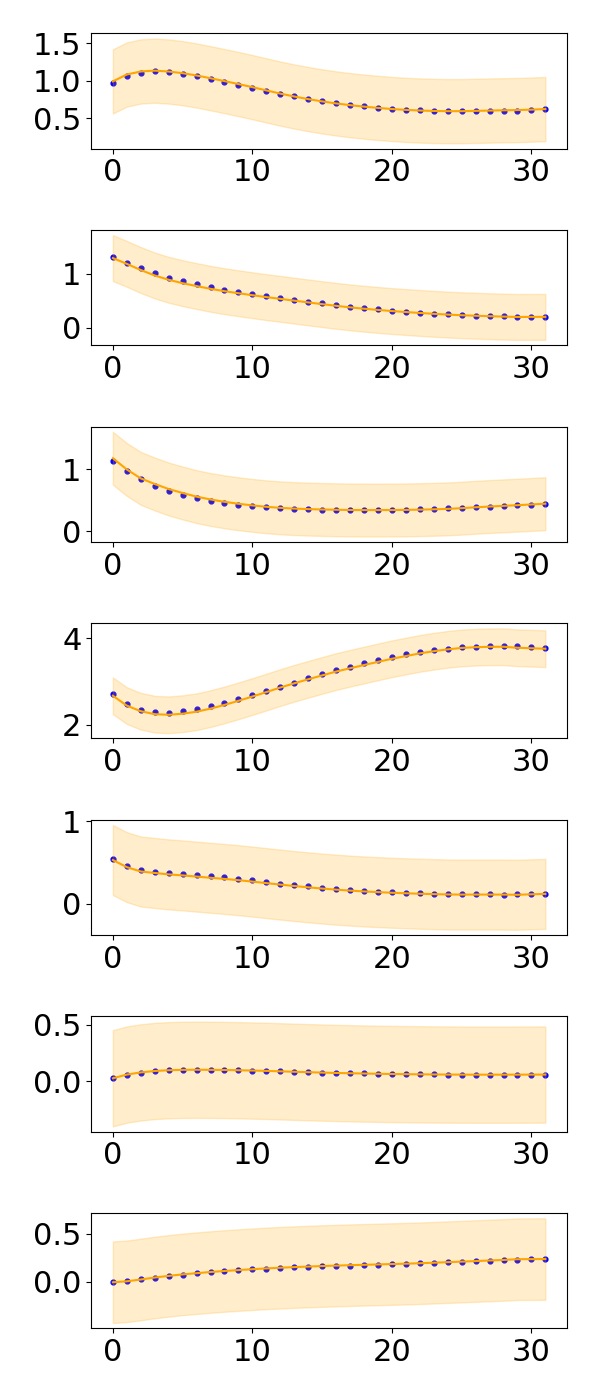}
    \includegraphics[scale=0.24]{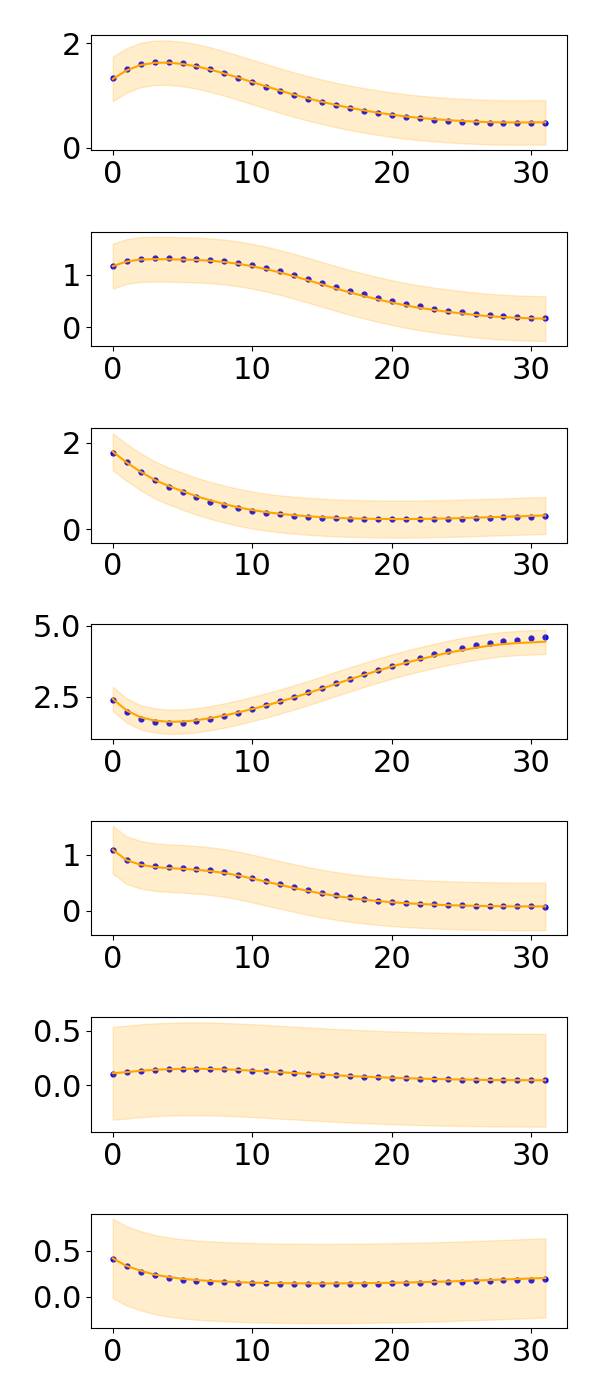}
    \caption{LALO:\textbf{ Neural SE}}
    \includegraphics[scale=0.24]{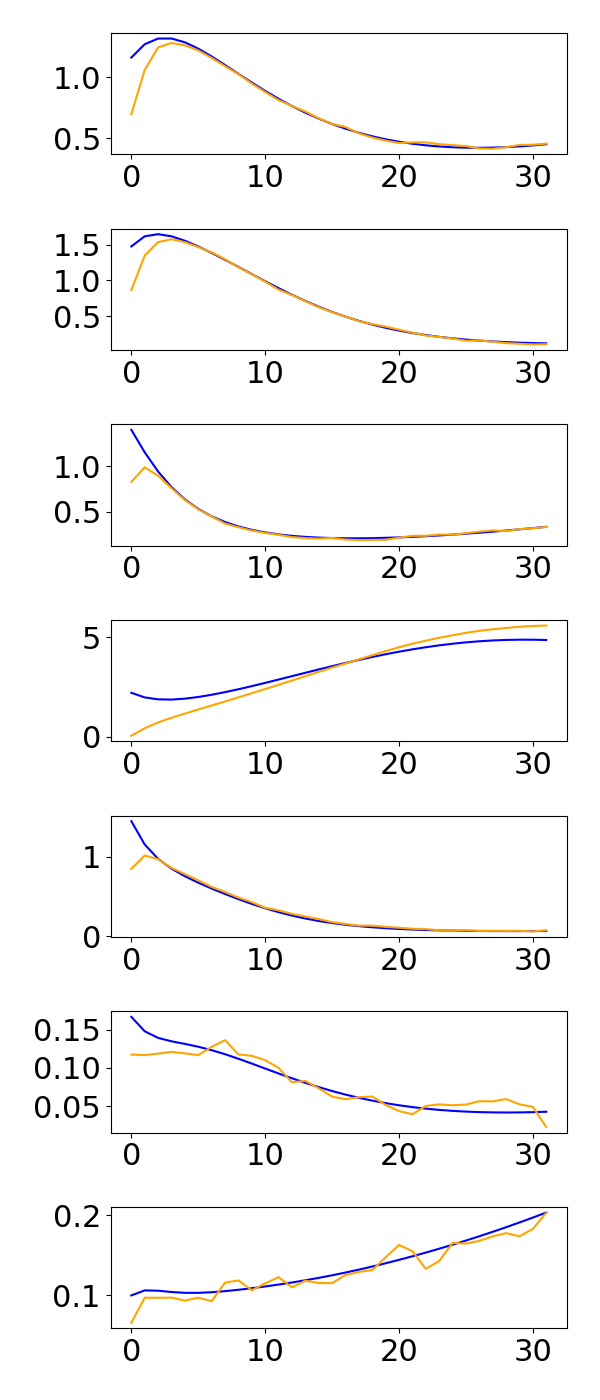}
    \includegraphics[scale=0.24]{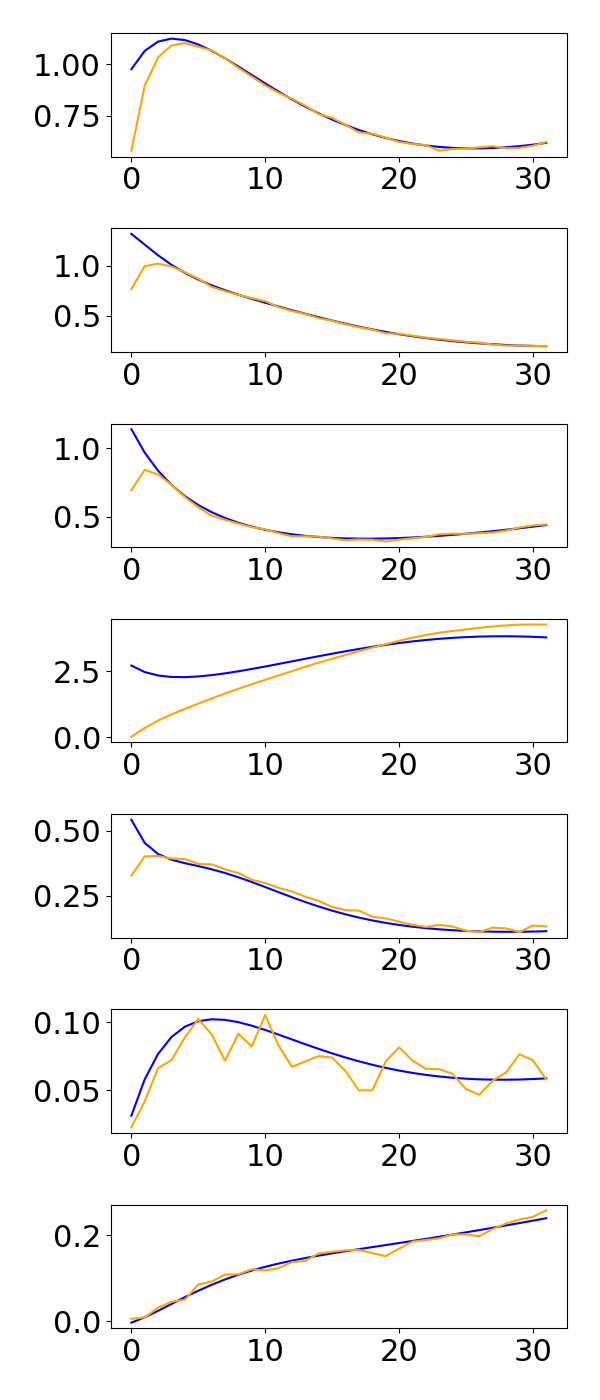}
    \includegraphics[scale=0.24]{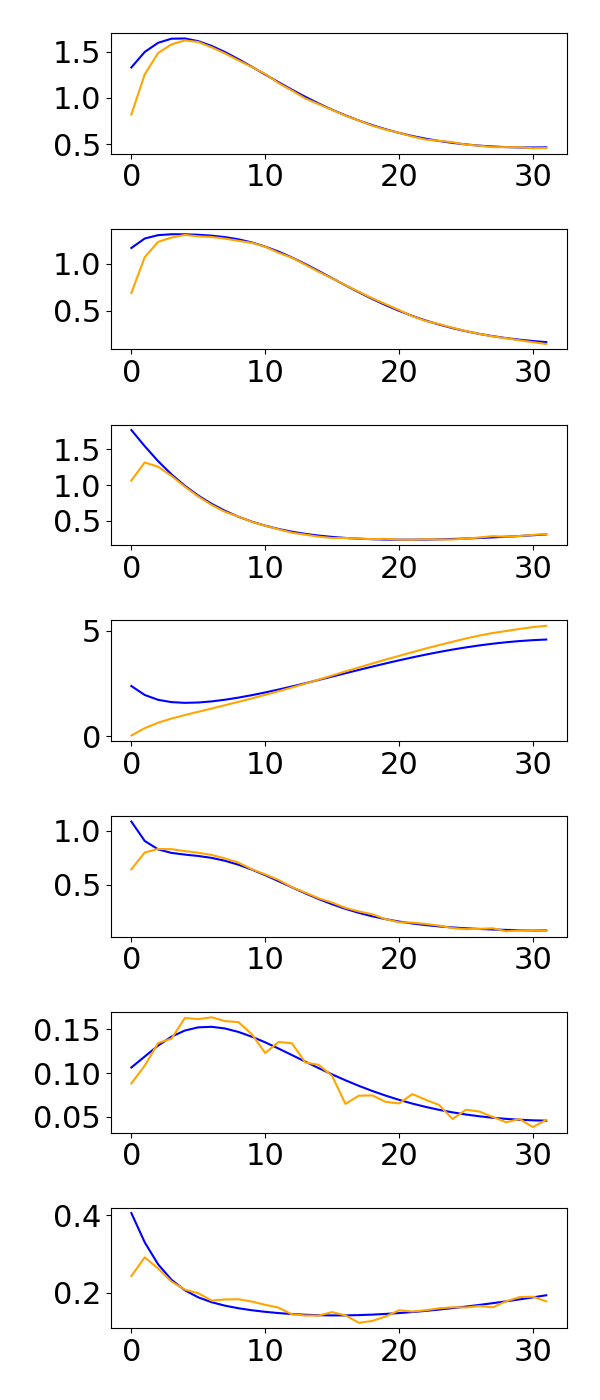}
    \caption{LALO: \textbf{Unscented Kalman Filters}. Each column is a different test point and each row is a variable of the state space.}
\end{figure}
\begin{figure}[ht]
    \centering

    \includegraphics[scale=0.23]{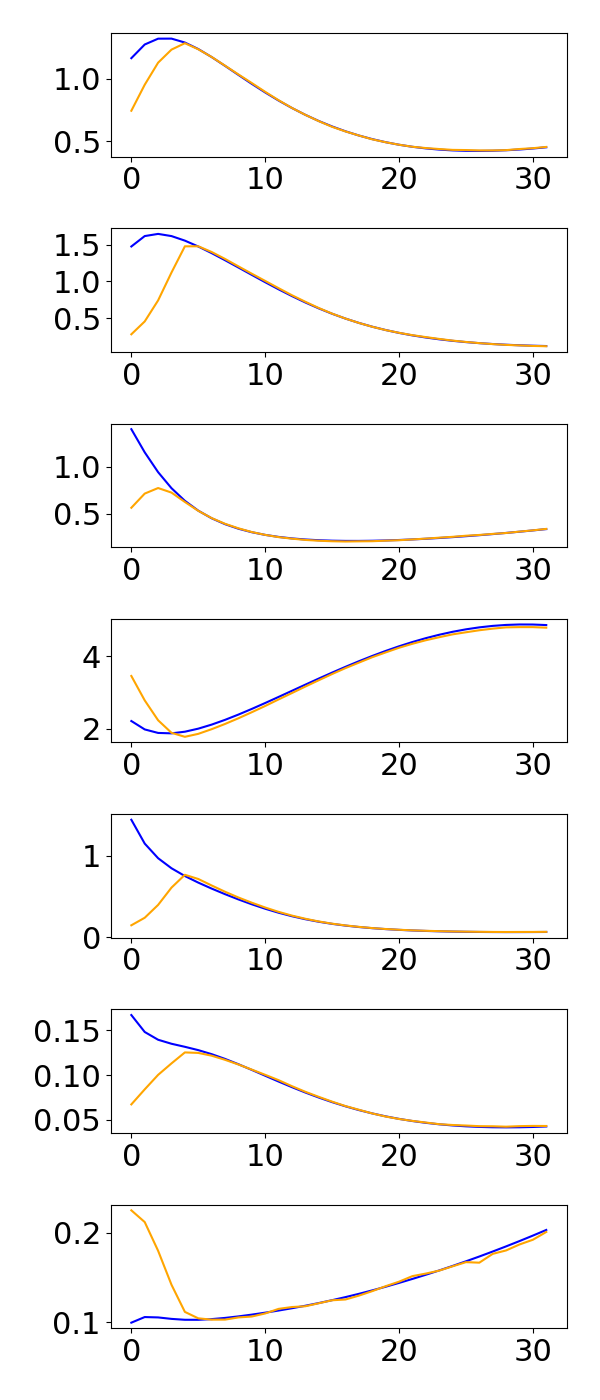}
    \includegraphics[scale=0.23]{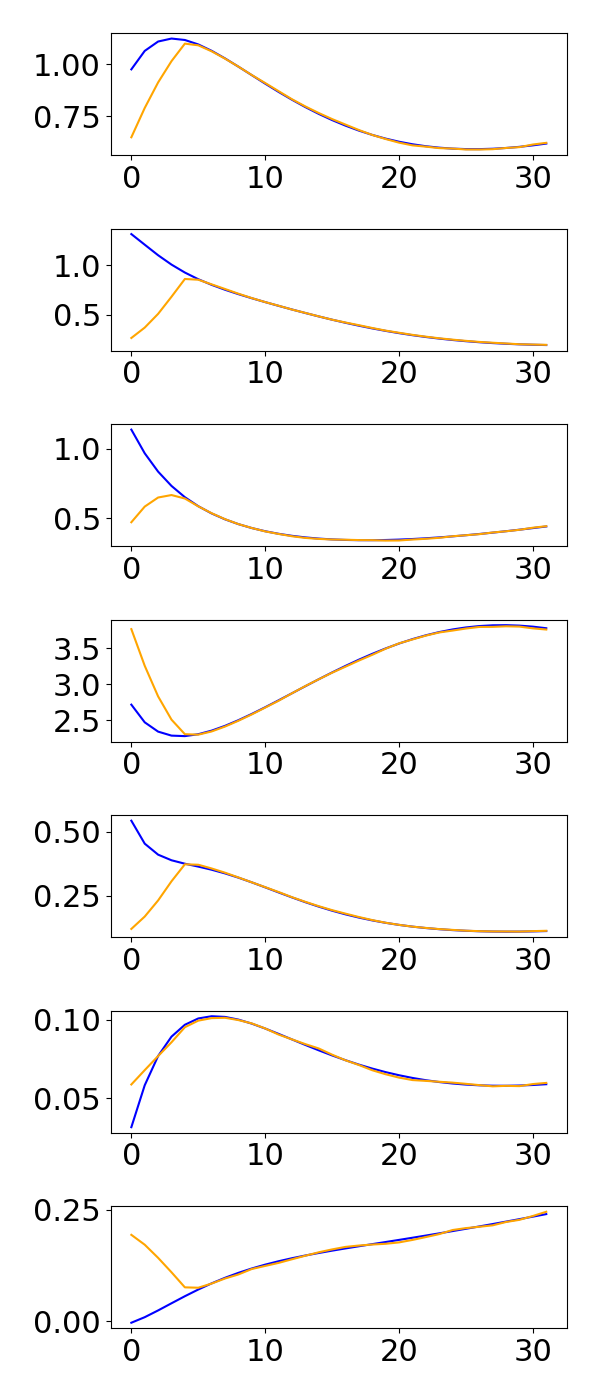}
    \includegraphics[scale=0.23]{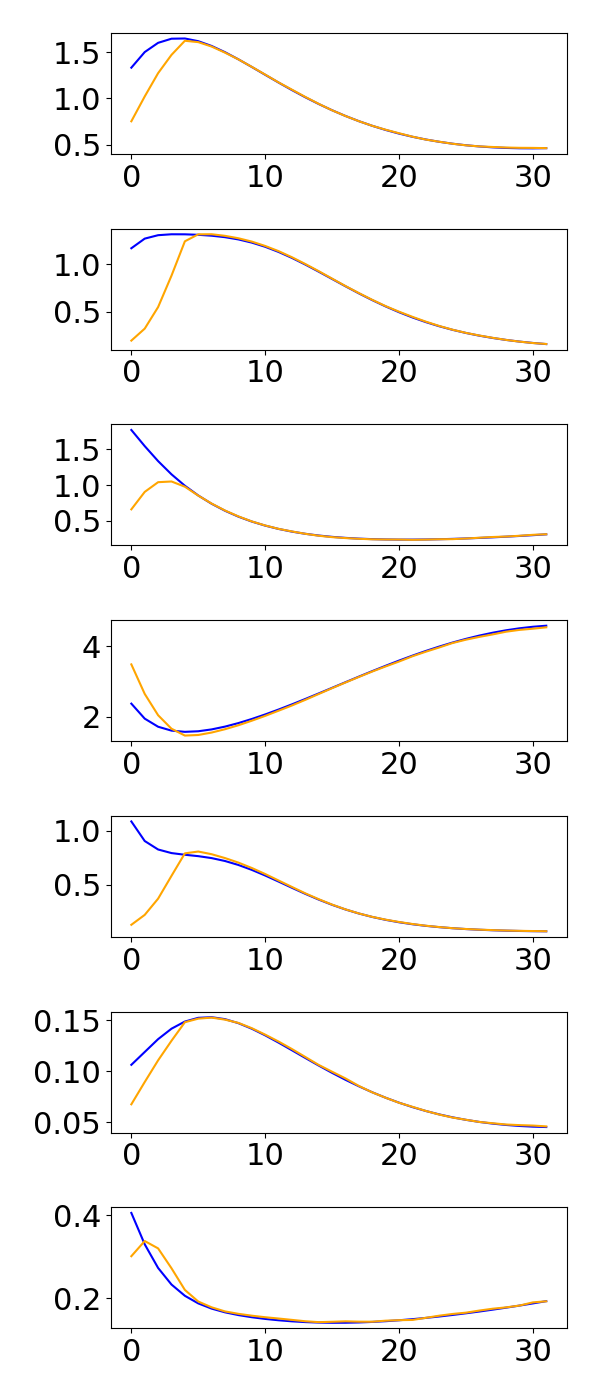}
    \caption{LALO: \textbf{Moving Horizon Estimate}. Each column is a different test point and each row is a variable of the state space.}
    \label{fig:LALO_SE}

    \includegraphics[scale=0.23]{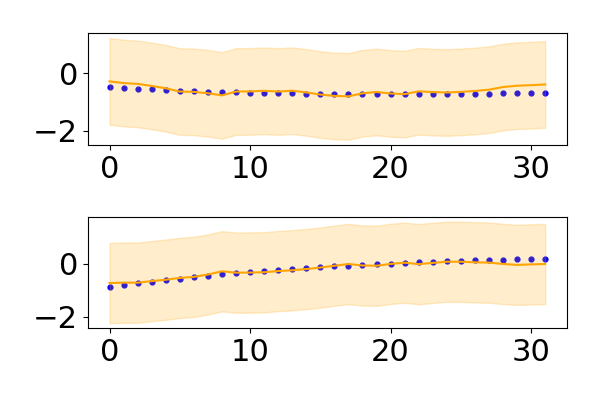}
    \includegraphics[scale=0.23]{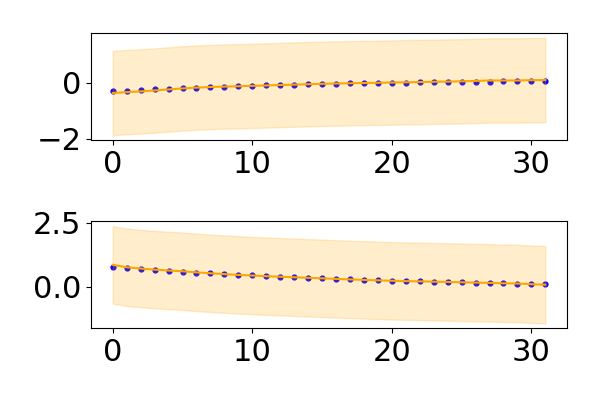}
    \includegraphics[scale=0.23]{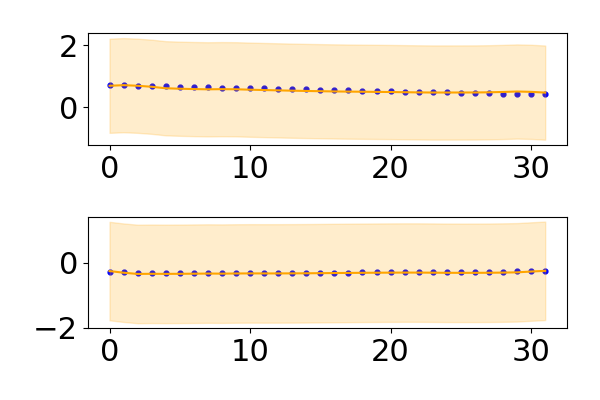}
    \caption{IP: \textbf{Neural SE}}
    \includegraphics[scale=0.23]{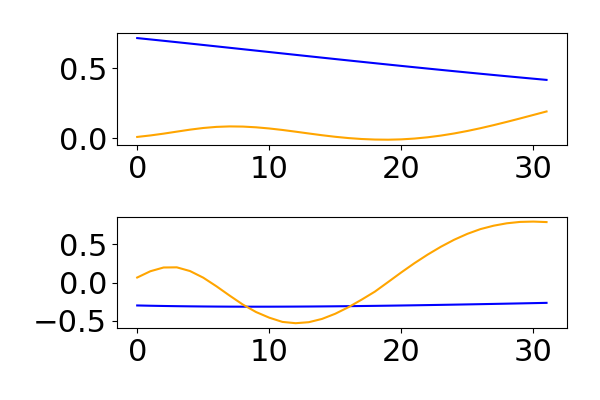}
    \includegraphics[scale=0.23]{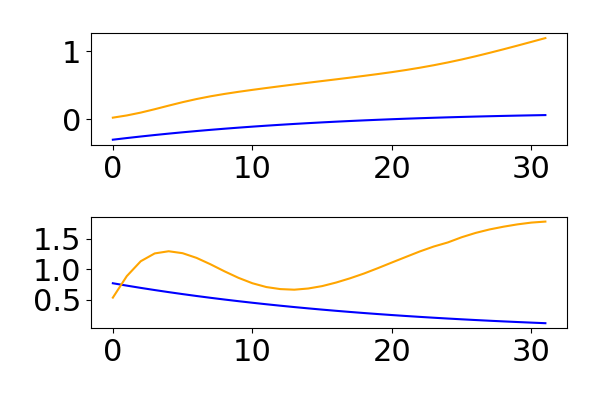}
    \includegraphics[scale=0.23]{se_imgs/IP3_smoothed_ukf_val_point_8.png}
    \caption{IP: \textbf{Unscented Kalman Filters}}
    \includegraphics[scale=0.23]{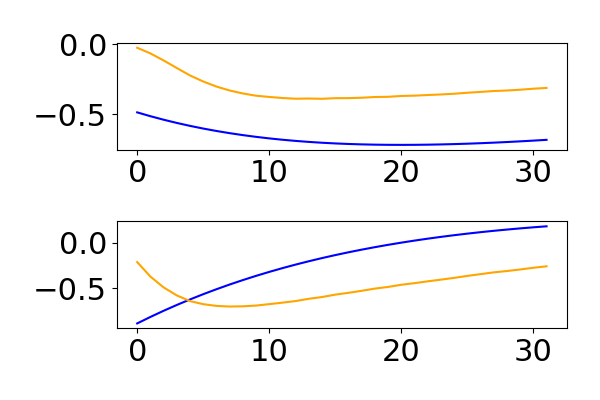}
    \includegraphics[scale=0.23]{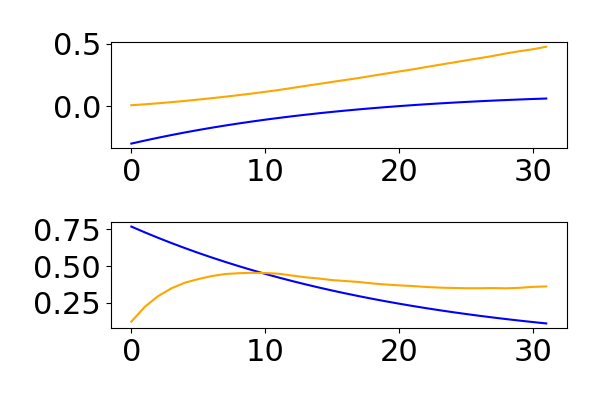}
    \includegraphics[scale=0.23]{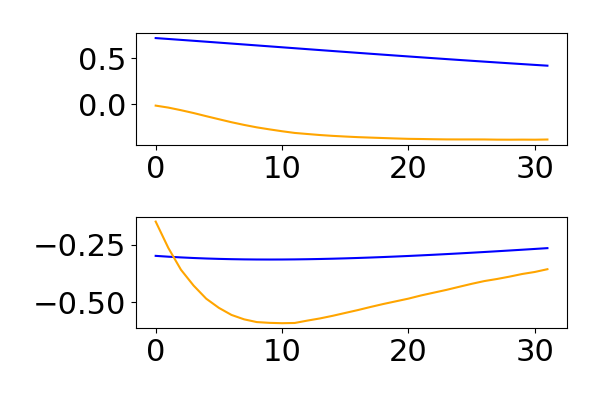}
    \caption{IP: \textbf{Moving Horizon Estimate}. Each column is a different test point and each row is a variable of the state space.}
    \label{fig:IP_SE}
\end{figure}

\begin{figure}[ht]
    \centering
\includegraphics[scale=0.24]{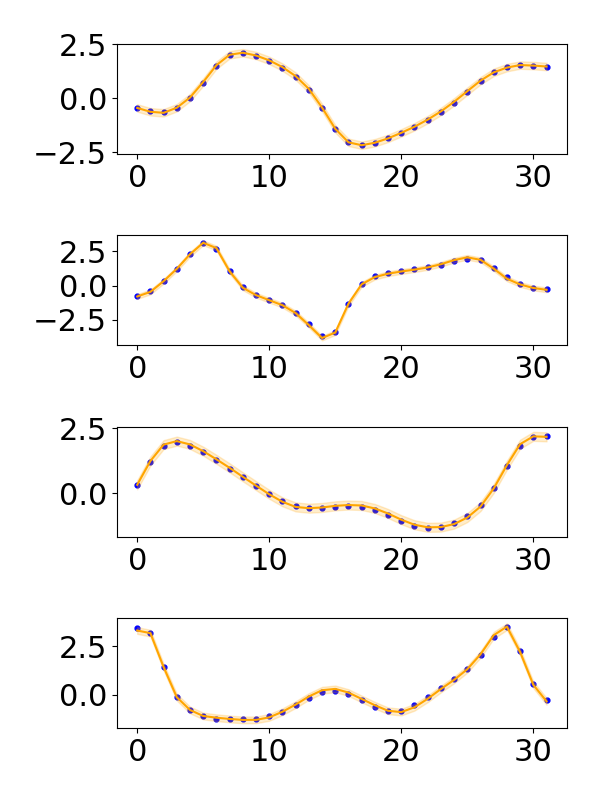}
    \includegraphics[scale=0.24]{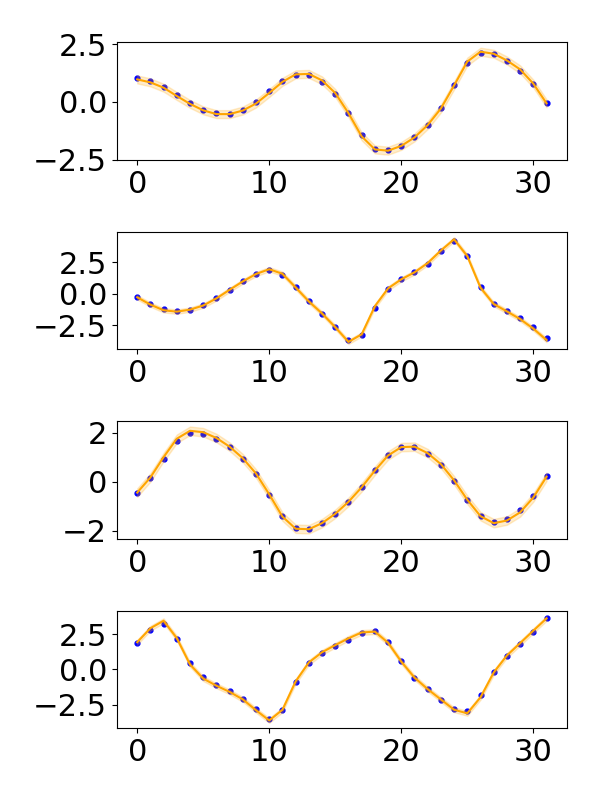}
     \includegraphics[scale=0.24]{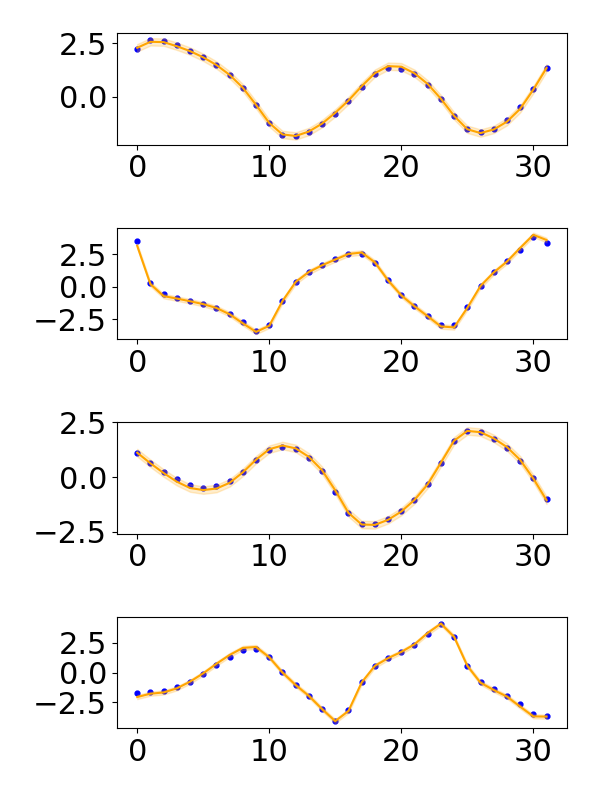}
    \caption{CVDP: \textbf{Neural SE}. Each column is a different test point and each row is a variable of the state space.}
    \includegraphics[scale=0.24]{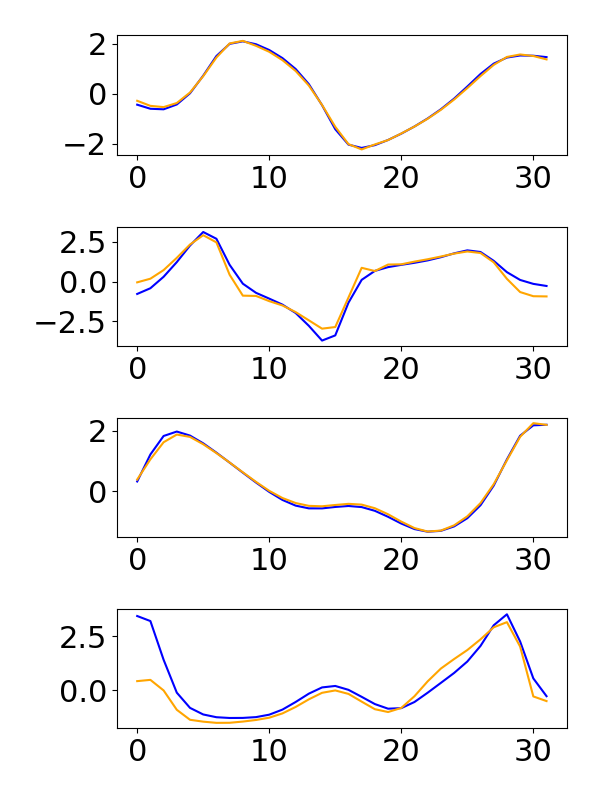}
    \includegraphics[scale=0.24]{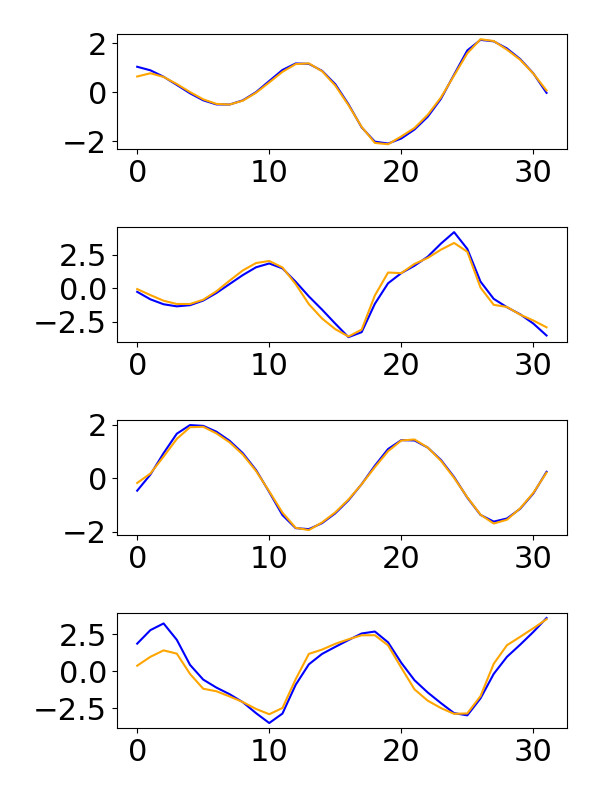}
        \includegraphics[scale=0.24]{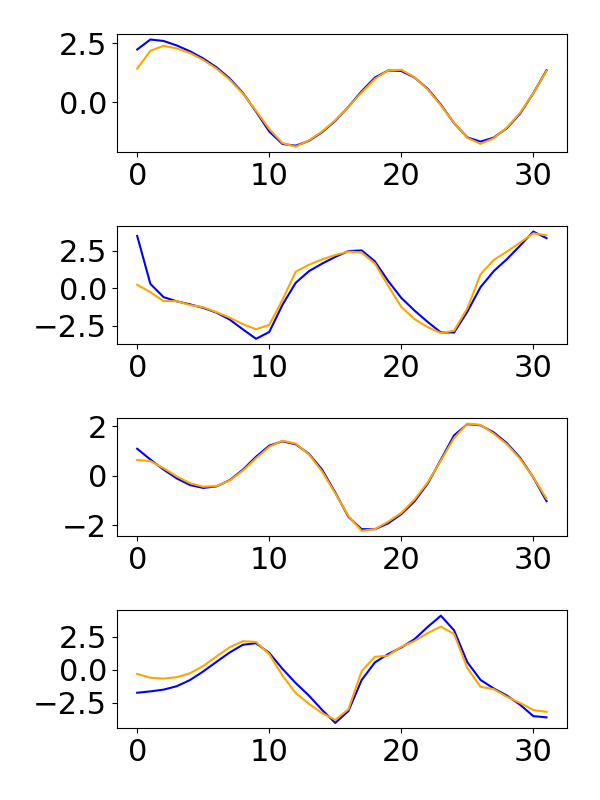}
    \caption{CVDP: \textbf{Unscented Kalman Filters}. Each column is a different test point and each row is a variable of the state space.}
    \includegraphics[scale=0.24]{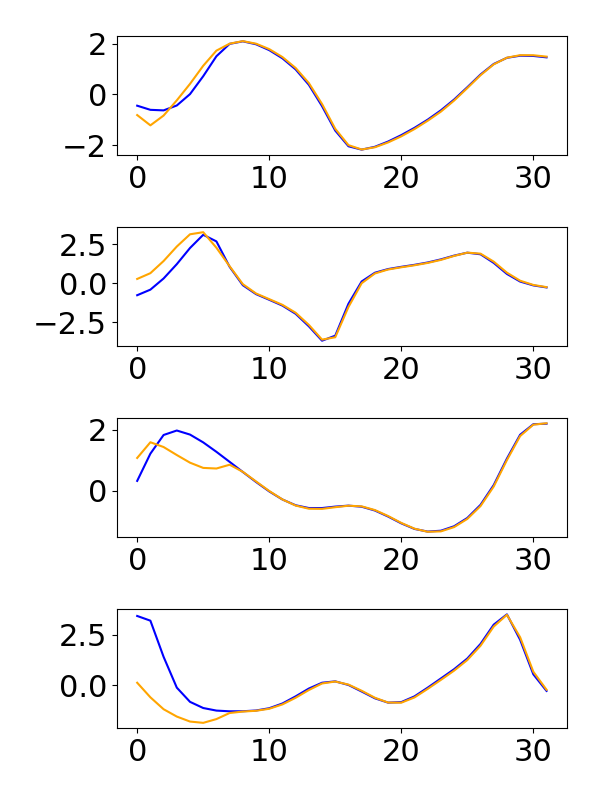}
    \includegraphics[scale=0.24]{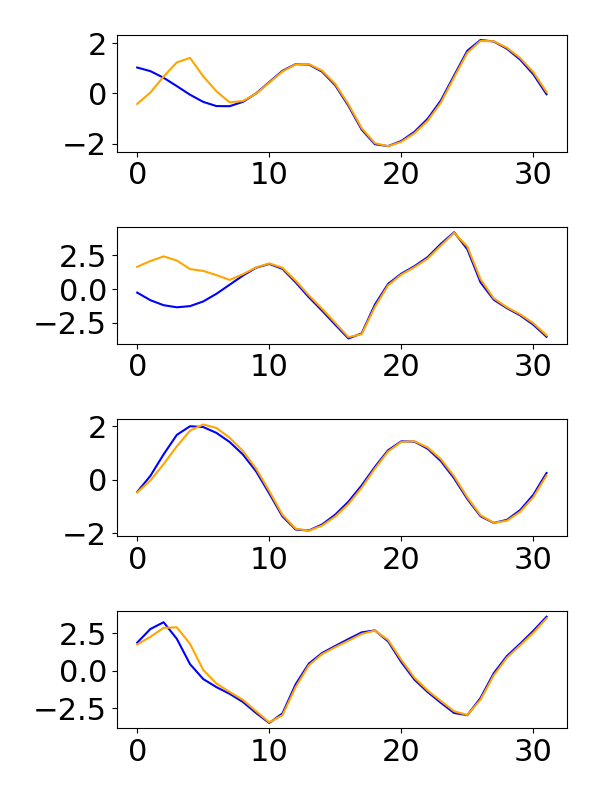}
    \includegraphics[scale=0.24]{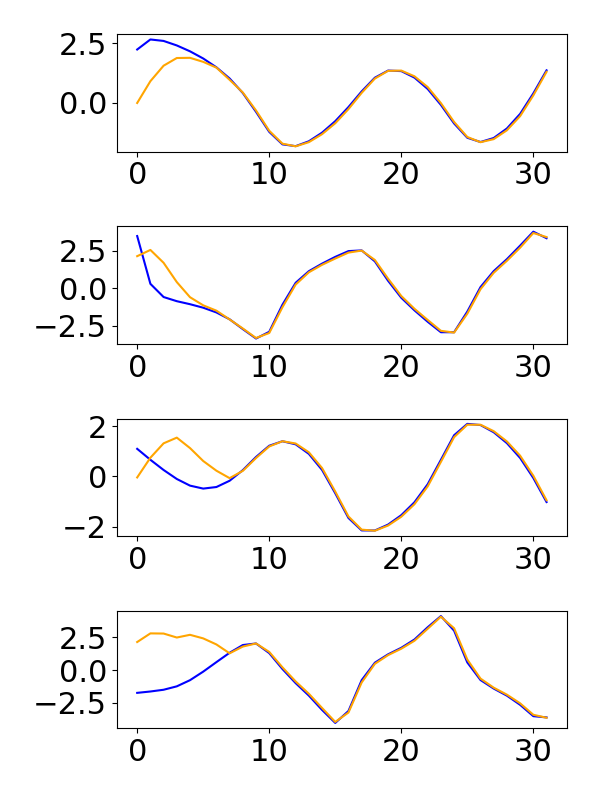}
    \caption{CVDP: \textbf{Moving Horizon Estimate}. Each column is a different test point and each row is a variable of the state space.}
    \label{fig:SE}
\end{figure}

\begin{figure}
    \centering
    \includegraphics[scale = 0.125]{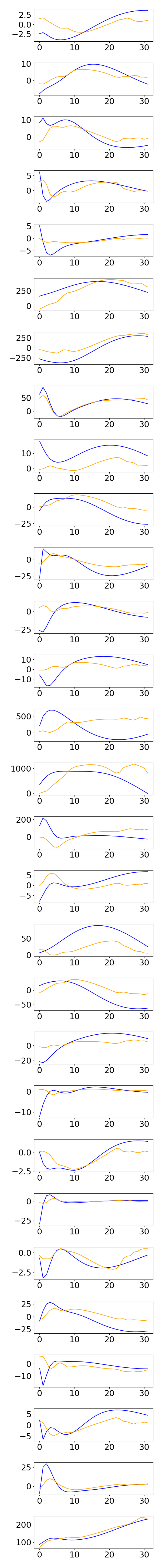}
    \includegraphics[scale=0.125]{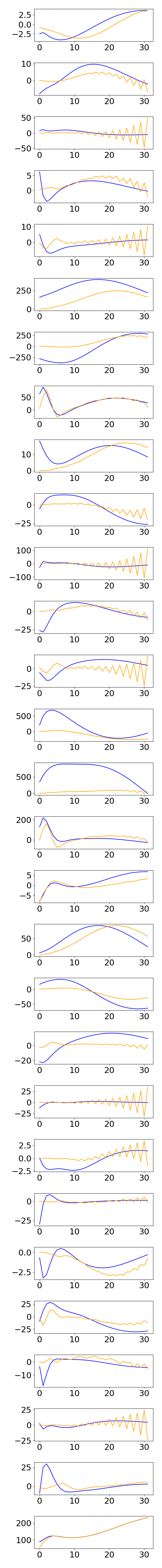}
    \includegraphics[scale = 0.125]{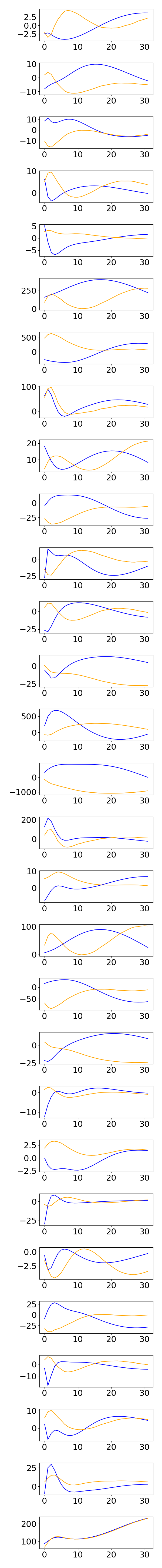}
    \caption{Helicopter: \textbf{Neural SE} vs \textbf{UKF} vs \textbf{MHE}}
    \label{fig:HC_SE}
\end{figure}

\begin{table}
\centering
\begin{tabular}{ |c|c|c|c| }
\hline
\textbf{Model} & \textbf{Neural SE} & \textbf{UKH} & \textbf{MHE}\\
\hline
\textbf{SN} & $0.0119\pm 0.0233$  & $0.5522\pm 0.5656$ & $0.7139\pm 0.7442$ \\
\textbf{IP} & $0.0233\pm 0.0401$
 & $0.1987\pm 0.1285$ & $0.1397\pm 0.1344$ \\
\textbf{CVDP} & $0.0040\pm 0.0039$  & $0.0210\pm 0.0321$ & $0.0337\pm 0.0763$\\
\textbf{TWT} & $0.0093\pm 0.0097$ & $0.0316\pm 0.0929$ & $1.3285\pm 0.2032$\\
\textbf{LALO} & $0.0081\pm 0.0071$& $0.0348\pm 0.0709$ & $0.0351\pm 0.1023$\\
\textbf{HC} & $0.0559\pm 0.0605$& $0.0832\pm 0.1065$ &$0.1217\pm0.1363$ \\
\hline
\end{tabular}
\caption{Comparison of the relative errors (mean and standard deviation over the test set) of the state estimators: the NeuralSE is compared to a Unscented Kalman Filter (UKF) and a Moving Horizon Estimator (MHE).\vspace{-0.5cm}}\label{table:se_comp}
\end{table}

\end{document}